\documentclass[lettersize,journal]{IEEEtran}
\usepackage{amsmath,amsfonts}
\usepackage{algorithmic}
\usepackage{algorithm}
\usepackage{array}
\usepackage[caption=false,font=normalsize,labelfont=sf,textfont=sf]{subfig}
\usepackage{textcomp}
\usepackage{stfloats}
\usepackage{url}
\usepackage{verbatim}
\usepackage{graphicx}
\usepackage{rotating}
\usepackage{multirow}
\usepackage{cite}
\usepackage{tabularx}
\usepackage{booktabs}

\begin{document}

\title{Combating Noisy Labels in \\ Long-Tailed Image Classification}

\author{ Chaowei Fang, Lechao Cheng, Yining Mao, Huiyan Qi, and  Dingwen Zhang~\IEEEmembership{Member,~IEEE,}
\thanks{Manuscript received June 6, 2022. \textit{(Corresponding Authors: Lechao Cheng, Dingwen Zhang.)}}
\thanks{Chaowei Fang is with School of Artificial Intelligence, Xidian University, Xi’an, 710071, China, and is also with Zhejiang Lab, Hangzhou, 310012, China. (e-mail:chaoweifang@outlook.com)}
\thanks{Lechao Cheng is with Zhejiang Lab, Hangzhou, 310012, China. (e-mail:chenglc@zhejianglab.com)}
\thanks{Yining Mao is with Zhejiang University, Hangzhou, 310012, China. (e-mail: yining.mao@zju.edu.cn)}
\thanks{Huiyan Qi is with Fudan University, Shanghai, 200433, China.}
\thanks{Dingwen Zhang is with Brain and Artificial Intelligence Laboratory, School of Automation, Northwestern Polytechnical University, Xi'an, 710072, China. (e-mail: zhangdingwen2006yyy@gmail.com)}

}



\maketitle

\begin{abstract}
Most existing methods that cope with noisy labels usually assume that the class distributions are well balanced, which has insufficient capacity to deal with the practical scenarios where training samples have imbalanced distributions. To this end, this paper makes an early effort to tackle the image classification task with both long-tailed distribution and label noise. Existing noise-robust learning methods cannot work in this scenario as it is challenging to differentiate noisy samples from clean samples of tail classes. To deal with this problem, we propose a new learning paradigm based on matching between inferences on weak and strong data augmentations to screen out noisy samples and introduce a leave-noise-out regularization to eliminate the effect of the recognized noisy samples. Furthermore, we incorporate a novel prediction penalty based on online prior distribution to avoid bias towards head classes. This mechanism has superiority in capturing the class fitting degree in real-time compared to the existing long-tail classification methods. Exhaustive experiments demonstrate that the proposed method outperforms state-of-the-art algorithms that address the distribution imbalance problem in long-tailed classification under noisy labels.
\end{abstract}

\section{Introduction}

\begin{figure}[t]
\centering
\includegraphics[width=\linewidth]{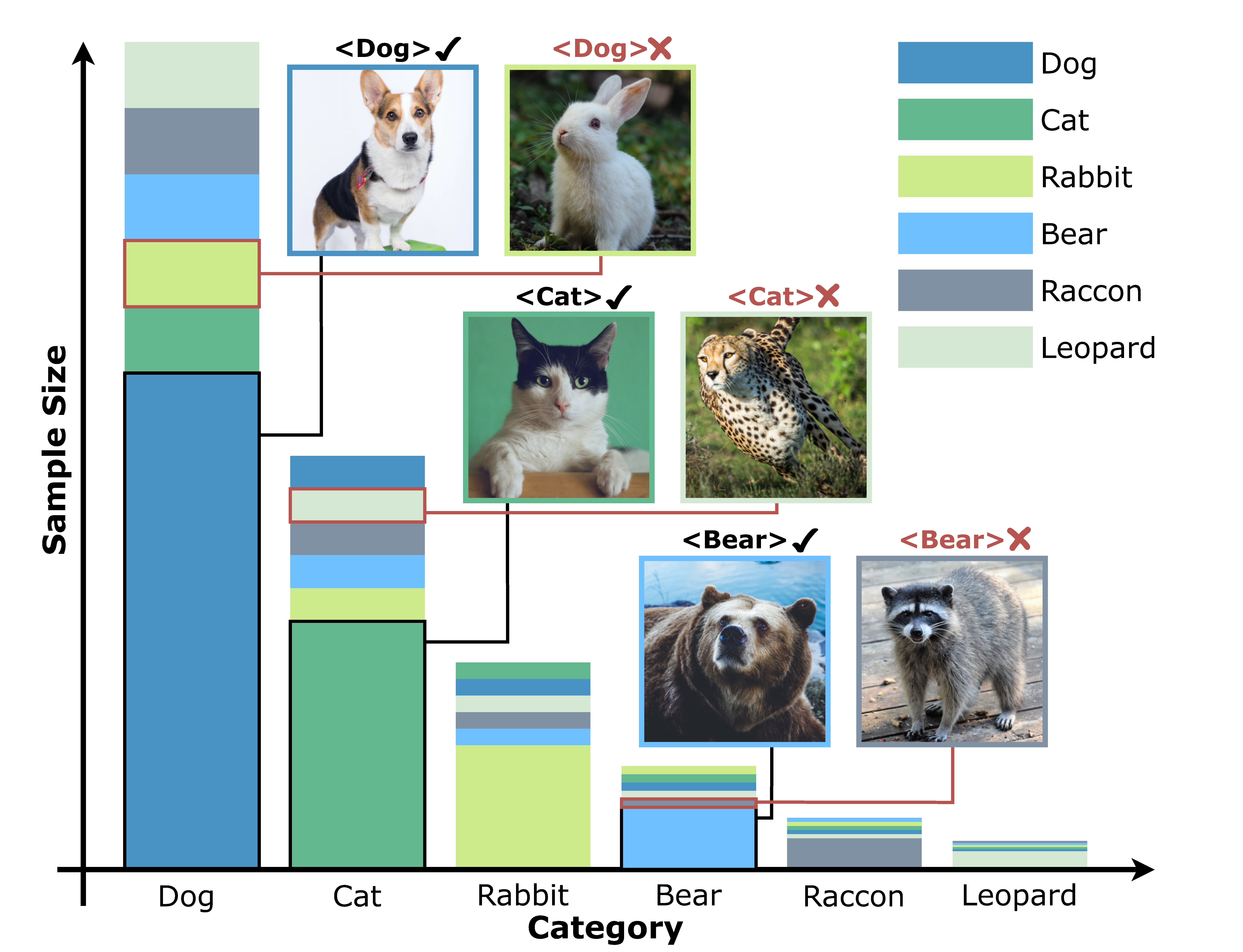}
\caption{This paper aims at tackling the learning task with both distribution imbalance and label noise.} \label{fig:teaser} 
\end{figure}

\IEEEPARstart{D}{eep} convolutional neural networks (CNNs) have achieved vast progress in addressing a variety of computer vision problems. 
Because CNNs have a large number of parameters, training them requires a massive demand for carefully labeled samples.
Researchers can utilize inexpensive labeling systems such as non-expert crowd-sourcing to reduce the cost of data annotation. However, these systems usually suffer from label noises.
Relieving the drawback of noisy annotations is crucial for CNNs under such circumstance.
Existing researches about learning with label noise typically focus on splitting clean and noisy examples while neglecting the distribution imbalance. Considering real-world data usually exhibits a long-tailed distribution, learning with both long-tailed distribution and label noise as in Fig.~\ref{fig:teaser} is urgent among real-world applications.

\begin{figure}
    \centering
     \subfloat[Cross Entropy]{%
       \includegraphics[width=0.48\linewidth]{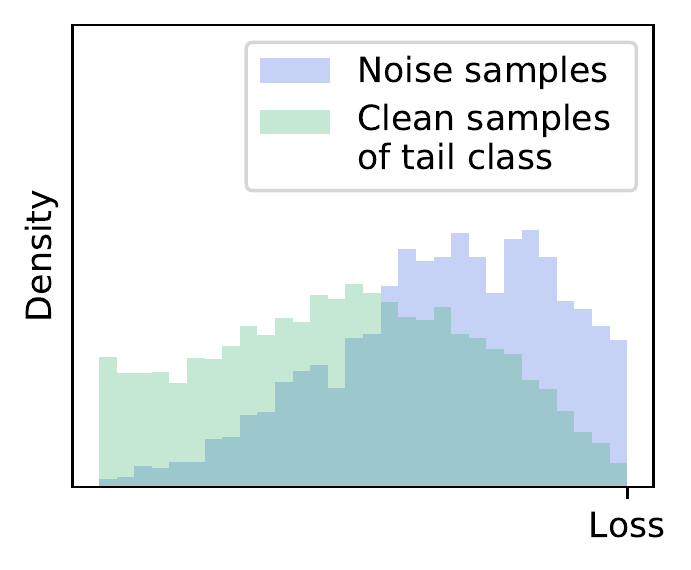}}
    \hfill
     \subfloat[Ours]{%
       \includegraphics[width=0.48\linewidth]{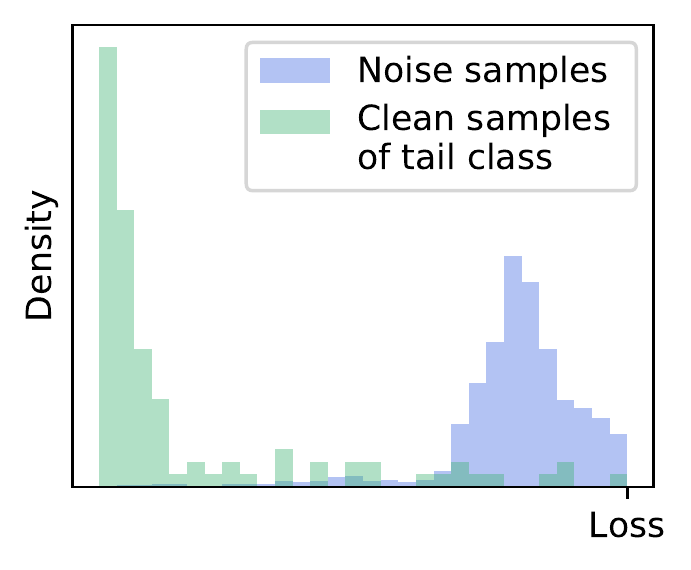}}    

\vspace{-0.5em}
\caption{Loss distributions of conventional cross entropy and our proposed cross-augmentation matching criterion. The loss values are calculated with ResNet32~\cite{he2016deep} on CIFAR-10. Labels of 20\% training images are modified. The ratio between the sample sizes of majority class and minority class is 100. } \label{fig:obs} 
\end{figure}

Deep CNNs have the capacity to memorize wrongly labeled samples during training but suffer from alarmingly increased generalization error~\cite{zhang2016understanding}. Thus, learning with label noise (LLN) attracts a large amount of research interest. 
\cite{arpit2017closer} reveals that CNNs first learn the simple and general pattern of clean samples before fitting noisy examples. Based on this concept, a variety of methods are proposed to mitigate the influence of noisy samples via modeling the cross-entropy loss between predictions and given labels~\cite{han2018co,yu2019does,nishi2021augmentation}. Samples with relatively smaller loss values are regarded as clean data and preserved for training models, while the remaining samples are excluded.  
However, under the long-tailed distribution, the cross-entropy loses between predictions and given labels are difficult to differentiate noisy samples from clean samples of tail classes, since the training data is mainly occupied by samples of head classes. As shown in Fig.~\ref{fig:obs} (a), we can observe this phenomenon on the CIFAR-10 dataset~\cite{2009Learning} with synthetic label noise.

Although the CNN model is capable of memorizing all training samples, noisy samples are more scattered than samples of tail classes, which means that conquering the former requires more effort.
Thus, predictions on noisy samples are more sensitive to strong image augmentation compared to predictions on clean samples of tail classes. Hence, we devise a specific cross-augmentation matching mechanism for recognizing noisy samples. 
As illustrated in Fig.~\ref{fig:obs} (b), the matching between predictions on weakly and strongly augmented images significantly improves the discrimination between noise samples and clean samples of the tail class.

 
In practice, we employ a shared network architecture for processing both weakly and strongly augmented images. However, strongly augmented images bias the feature distribution, causing shifts on the statistics of the batch normalization. Therefore, we devise a dual-branch batch normalization module to process weakly and strongly augmented images with separate parameters. 
Based on the cross-augmentation matching, noisy samples can be well differentiated from clean samples under both distribution imbalance and label noise.
For re-balancing clean samples, we propose to penalize the prediction with online prior category probabilities. 
The class-wise prior probability is dynamically estimated from online predictions of the classification model.
Compared to existing strategies such as re-sampling~\cite{more2016survey} and re-weighting~\cite{cui2018large}, which depend on sample sizes of classes to re-balance the training data,
the online prior penalization is robust against the existence of noisy samples in which actual class sample sizes are unavailable. The other advantage of our method is that it can reflect the practical fitting degree of the classification model on classes. 
For combating the recognized noisy samples, a leave-noise-out regularization is devised by suppressing predictions on labels of the recognized noisy samples.
Experiments on five public benchmarks indicate that our proposed pipeline significantly outperforms existing methods for learning with label noise and long-tailed distribution.   

The main contributions of this paper are as follows:
\begin{itemize}
    \item [1)] We propose a novel method that can effectively detect noisy samples under long-tailed distribution, based on matching between predictions on weak and strong data augmentations. Besides, a leave-noise-out regularization is introduced to counteract the impact of noisy samples.
    \item [2)] The prior distribution is dynamically evaluated for rebalancing the training data.
    \item [3)]  Exhaustive experiments on five public datasets are conducted to validate the superiority of our method compared to existing methods.
\end{itemize}

\section{Related Work}
\subsection{Deep Learning with Long-Tailed Distribution}
In~\cite{zhang2021bag}, a detailed recipe is provided to guide the choice of techniques effective in long-tailed classification, such as re-sampling, re-weighting, and mixup~\cite{zhang2017mixup}.
However, the re-sampling/re-weighting methods harm the learning of feature representations~\cite{zhou2020bbn}. 
Another popular solution is formed by a two-stage pipeline, including representation learning based on uniform sampling and classifier learning based on re-balanced sampling~\cite{cui2018large,cao2019learning}. 
In~\cite{zhou2020bbn}, a cumulative learning algorithm is devised to combine the advantages of the representation capacity of the model trained with conventional uniform sampling and the classification capacity of the model trained with inverse proportional sampling.
\cite{zhong2021improving} designs hand-crafted functions for smoothing classes concerning their sample sizes in the training dataset.  
Few studies pay attention to task of learning with both distribution imbalance and noisy label. \cite{zhong2019unequal} attempts to settle the face recognition task in such a setting. However, this kind of study on general image classification is still under-explored.
We make an early effort to tackle the learning task with both long-tailed distribution and label noise on general image classification. 
Furthermore, we propose a novel prior penalization term to smooth the labels of recognized clean samples according to the real-time confidence levels on classes. Compared with existing data rebalancing strategies which rely on class sample sizes, our devised mechanism is more robust to label noise, and can capture the real-time fitting degree of classes. 

\subsection{Deep Learning with Label Noise}
The key to solving the learning problem with label noise is filtering out noisy samples before they are memorized by the model. One type of algorithms is devised on the basis of the co-training mechanism, such as \cite{han2018co,yu2019does,li2020dividemix,lu2021co}, which unify two classification models to prevent them from memorizing noisy samples.
\cite{han2018co,yu2019does} employs two models to select small-loss samples as clean samples for each other.
\cite{li2020dividemix} incorporates the mixup operation
to rectify the noisy labels with the help of predictions from peer networks.
Instead of maintaining two diversified models, 
\cite{wei2020combating} attempts to encourage the agreement between two models and devise a co-regularization mechanism to select clean samples.
\cite{lu2021co} builds up two models for weakly and strongly augmented images respectively and adopts the consistency loss between their predictions to preclude noisy samples.
\cite{yao2021jo,yi2021transform} also detect noisy samples according to the deviation between predictions on differently augmented images.  
\cite{yi2021transform} employs a single network to cope with weakly and strongly augmented images simultaneously. However, the strongly augmented images interfere the feature distribution captured on the training data.
\cite{yao2021jo} leverages the mean teacher model to rectify labels of noisy samples. However, the rectified labels may be incorrect, especially in early training stages. 
In this paper, we aim to settle the task of learning with noisy labels under severe long-tailed distribution.
Inspired from~\cite{xie2019unsupervised,sohn2020fixmatch,lu2021co}, the matching between predictions on weakly and strongly augmented images is utilized to screen out noisy samples considering it can well separate noisy samples from clean samples of tail classes.
Instead of using two networks~\cite{lu2021co}, we construct only one network with the help of the dual-branch batch normalization, which can avoid the influence of strongly augmented images on feature distribution. Our method is more economical and capable of leveraging strongly augmented images to facilitate representation learning without introducing extra side effects.
The main differences between our method and existing LLN methods are as follows.
1) Previous works are limited to conquering noisy labels under balanced data distribution, while our method takes the data imbalance into consideration.
2) We devise a novel method for preventing learning bias towards head classes via online prior penalization. A leave-noise-out regularization is devised to erase the negative effect caused by noisy samples.
Online label smoothing~\cite{zhang2021delving} is targeted at tackling the overfitting problem in image classification without data imbalance.  It helps to prevent over-confidence on target categories and explore underlying similarity knowledge among different categories. In our method, the online prior category distribution is used for relieving the learning bias towards head classes via penalizing predictions on them.



\section{Method}
\subsection{Preliminary}
This paper aims at tackling the long-tailed classification under the existence of noisy labels.
Given a training dataset $\mathbb D=\{(x_i,y_i)\}_{i=1}^N$ with $N$ samples and $K$ classes, we learn a deep CNN model to implement the classification task. Here, $y_i$ ($\in\{1,2,\cdots,K\}$) is possibly incorrect.  Denote the number of samples in the $k$-th class be $N_k$. In many practical applications, samples are severely imbalanced with respect to classes. 

\subsection{Observation}
In the noisy label learning task under long-tailed distribution, the training data is dominated by samples of head classes. The model merely captures patterns of head classes in the early stage. As shown in Fig.~\ref{fig:obs} (a), only using the cross-entropy loss between predictions and given labels, which is
widely applied to detect noisy samples e.g.~\cite{han2018co,li2020dividemix}, is hard to discriminate samples with wrong labels from those of the tail class. There exists a large overlap between losses on the two kinds of samples.

\subsection{Overview}
\begin{figure*}[htp]
\centering
\vspace{-0.5em}
\includegraphics[clip, trim=0.2cm 0cm 0.4cm 0.2cm,width=0.95\textwidth]{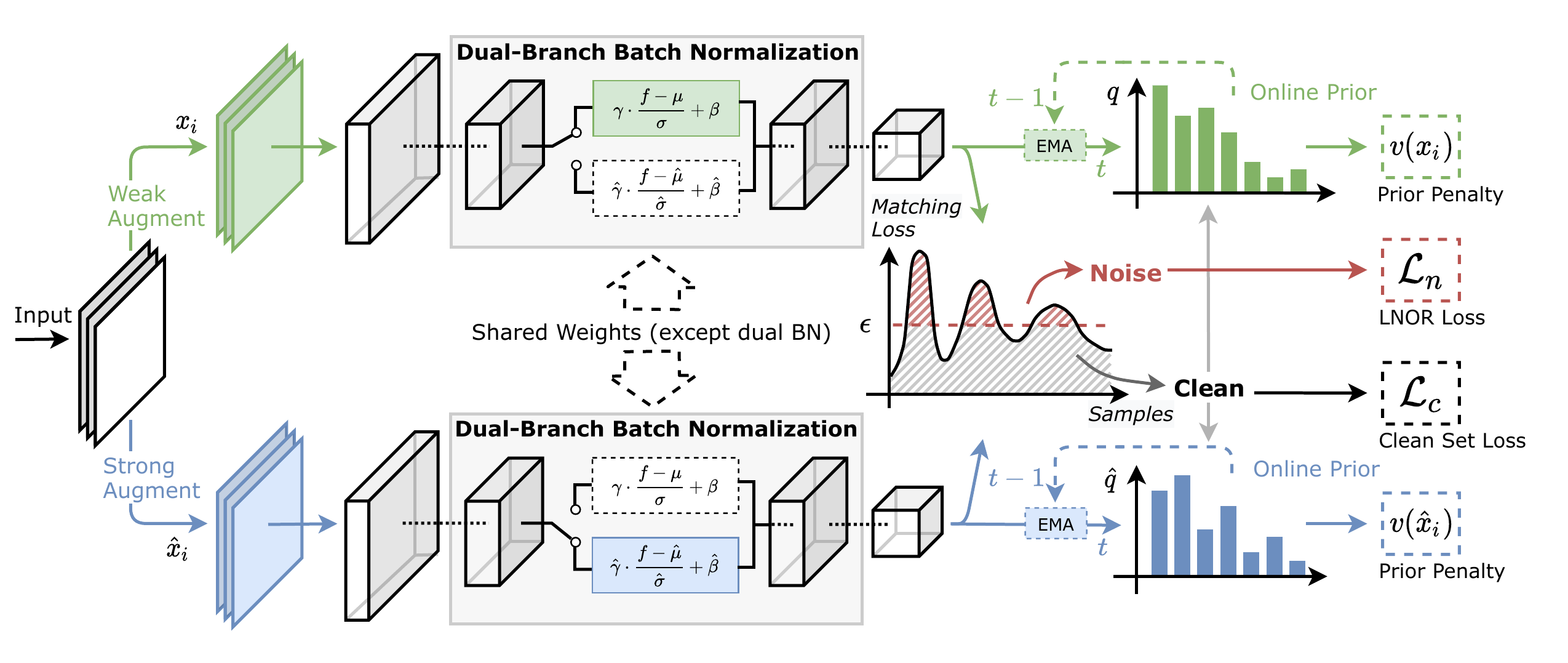}
\caption{Overview of our proposed method. 
The matching between predictions on weakly and strongly augmented images is involved in to identify clean and noisy samples. Leave-noise-out regularization (LNOR) is employed to erase the impact of the recognized noisy samples, and an online prior penalization scheme is devised to re-balance training data. `EMA' indicates the exponential moving average. } 
\label{fig:framework}  
\end{figure*}

We propose a novel algorithm to solve image classification with both distribution imbalance and label noise. 
The CNN-based classification model is built up with the help of the dual-path normalization, which processes feature maps of weakly and strongly augmented images with separate parameters. Based on it, we employ the cross-augmentation matching to detect noisy labels, which effectively tackles the issue in the above observation. 
When learning on the clean set of samples, an online prior penalization strategy is devised for balancing training data.
We propose to eliminate the effect caused by noisy samples through restraining predictions on  given labels of these recognized noisy samples. The overall framework of our method is illustrated in Fig.~\ref{fig:framework}.

\subsection{Cross-Augmentation Matching} \label{sec:caug}
We involve in the inference matching between weakly and strongly augmented images for splitting wrongly labeled samples and clean samples of tail classes.  Suppose the prediction of the classification model on $x_i$ be $h(x_i \mid \theta) \in \mathbb R^K$. $\theta$ indicates network parameters, and $h(\cdot)$ denotes the mapping function. The following function is used to judge whether $x_i$ is correctly labeled, 
\begin{align} 
\label{eq:loss-match} \ell(x_i,y_i) &=\ell^c(x_i,y_i)+ \ell^c(\hat x_i, y_i)+ \alpha \ell^c(\hat x_i, y_i^\prime), \\
\label{eq:loss-ce} \ell^c(x_i,y_i) &= -\mathbf y_i^T \log(h(x_i\mid\theta)), 
\end{align}
where $\alpha$ is a weighting coefficient. $x_i$ and $\hat x_i$ represents the weakly and strongly augmented images respectively. $y_i^\prime=\arg\min_k p_i[k]$, which denotes the most confident class predicted on $x_i$. $\mathbf y_i$ is the one-hot format of $y_i$. As demonstrated in Fig.~\ref{fig:obs} (b), the adoption the loss criterion in Eq.~\ref{eq:loss-match} significantly improves the distinguishability between noisy samples and clean samples of tail classes, in contrast to the pure cross entropy between predictions and given labels, namely $\ell^c(x_i,y_i)$.

Samples with losses smaller than the OTSU threshold $\epsilon$ are regarded as clean data, $\mathbb C=\{(x_i,y_i)\mid\ell(x_i,y_i)<\epsilon\}$. 
During the training stage, Eq.~\ref{eq:loss-match} is utilized to calculate losses for samples in $\mathbb C$. The overall objective function on $\mathbb C$ is formulated as,
\begin{equation} \label{eq:loss-clean}
    L_c = \sum_{(x_i,y_i)\in\mathbb C} \ell(x_i,y_i).
\end{equation}

\vspace{1mm}
\noindent \textbf{Dual-Branch Batch Normalization} 
Since strongly augmented images have evidently different appearances compared to original images, they will bring interference to the feature distribution captured by the CNN model.
Thus, when constructing the CNN model, we incorporate the dual-branch batch normalization to deal with the distribution deviation between original and strongly augmented images. 
Specific mean, variance and affine parameters are used to normalize features of weakly and strong augmented images respectively, inspired from~\cite{chang2019domain}. Suppose the feature generated by a convolution layer be $\mathbf f_i$ and $\hat{\mathbf f}_i$, given weakly augmented image $\mathbf x_i$ and the strongly augmented counterpart $\hat{\mathbf x}_i$. Distinct means and variances are calculated for the two types of features,
\begin{align}
    \mu = \frac{1}{m} \sum_{i=1}^m \mathbf f_i,\; \sigma=\sqrt{\frac{1}{m}\sum_{i=1}^m (\mathbf f_i-\mu)^2+\xi}; \\
    \hat{\mu} = \frac{1}{m} \sum_{i=1}^m \hat{\mathbf f}_i,\; \hat \sigma=\sqrt{\frac{1}{m}\sum_{i=1}^m (\hat{\mathbf f}_i-\hat{\mu})^2+\xi};
\end{align}
where $\xi$($=10^{-5}$) is a constant, and $m$ represents the number of sample in single batch. These statistical values are separately accumulated via exponential moving average. 
The outputs of the two branches in the batch normalization are denoted as, $\mathbf f_i^\prime = \gamma \frac{\mathbf f_i-\mu}{\sigma}+\beta$ and $\hat{\mathbf f}_i^\prime = \hat{\gamma} \frac{\hat{\mathbf f}_i-\hat{\mu}}{\hat{\sigma}}+\hat{\beta}$, where $\gamma$, $\beta$, $\hat{\gamma}$, and $\hat{\beta}$ are learnable affine parameters.  
During training, each kind of augmented images is forwarded with one group of parameters. 
During testing, input images are merely processed with simple transformations, having a more similar distribution to weakly augmented images.
Thus, only the parameters estimated from weakly augmented images are preserved while those estimated from strongly augmented images are deprecated.

\vspace{1mm}
\noindent \textbf{Leave-Noise-Out Regularization}
Samples with very large loss values are considered as confident noisy samples, which are denoted as, 
\begin{align}
    \nonumber &\mathbb N=\arg\max_{\mathbb S} \sum_{(x_i,y_i)\in \mathbb S} \ell(x_i,y_i),\\
    \label{eq:noisy} &\textrm{s.t.}\;\mathbb S\cap \mathbb C=\emptyset,\;\; \mid\mathbb S\mid \leq \kappa (\mid\mathbb D\mid-\mid\mathbb C \mid).
\end{align}
$\kappa\in(0,1)$ is a constant. Considering the given labels of these confident noisy samples are incorrect, they are used to regularize the network optimization via a specific leave-noise-out regularization. We can assume that each sample $x_j$ in $\mathbb N$ belongs to certain class from $\{k=1,\cdots,K\mid k\neq y_j\}$. The following regularization term is used to constrain network predictions on $\mathbb N$,
\begin{equation} \label{eq:loss-noisy}
    L_n = -\sum_{(x_j,y_j)\in\mathbb N} \log(1- \mathbf p_j[y_j]).
\end{equation}
The above regularization prevents predictions of these recognized noisy samples from being positive on their labels. It benefits to cancel out the negative impact of noisy samples.



\subsection{Online Prediction Bias} \label{sec:opp}
For relieve the dominance of head classes under the long-tailed distribution, existing methods focus on re-sampling~\cite{more2016survey,buda2018systematic,kang2019decoupling} or re-weighting~\cite{cui2018large} training data. However, these methods are dependent to real class sample sizes which are unavailable when label noise exists. Additionally, fixed class sample sizes can not reflect the practical fitting degree on classes.
To settle the above problems, we propose to smooth the predictions on samples of head classes by penalizing their confidences with the online prior probabilities. Practically, we choose to evaluate categorical prior probabilities from online predictions of the network, which can authentically reflect the fitting degree on classes. The prior probability of the $k$-th class is dynamically estimated as follows,
\begin{equation} \label{eq:prior}
\mathbf q := (1-\tau) \mathbf q +\tau \sum_{(x_i,y_i)\in\mathbb C}  h(x_i\mid\theta)/\mid\mathbb C\mid,
\end{equation}
where $\tau$ is a constant. $\mathbf q$ is initialized by the ratios of samples in $K$ classes, namely $q[k]=\frac{N_k}{N}$. 
The online prior penalization is defined as follows,
\begin{equation}
    \label{eq:penalty} v(x_i) = - (1-\mathbf q^T) \log(h(x_i\mid\theta)).
\end{equation}

We can acquire the following loss function after incorporating the penalty of prior probabilities into the cross-entropy function,
\begin{equation}\label{eq:loss-penalized}
    \ell_p(x_i,y_i) = \ell^c(x_i,y_i)+\lambda v(x_i),
\end{equation} 
where $\lambda$ is a constant weighting coefficient. 
It can be transformed into the following formulation,
\begin{align}
    \ell_p(x_i,y_i) &= -(1+\lambda K-\lambda) \sum_{k=1}^K s_i^k \log(p_i[k]), \\
    s_i^k &= \begin{cases}
    \frac{1+\lambda(1-q[k])}{1+\lambda K-\lambda},\;\;\; k=y_i, \\
    \frac{\lambda(1-q[k])}{1+\lambda K-\lambda},\;\;\;\;\;\;\; k\neq y_i.
    \end{cases}
\end{align}
We can observe that the penalty item in Eq.~\ref{eq:penalty} is able to smooth labels according to the prior distribution. Labels with larger prior probabilities are softened more intensively. Meanwhile, the reduction of the target label is distributed to other labels in negative relation to their probabilities. This helps to bias the optimization process towards tail classes. The online prior penalization is applied for predictions on both weakly and strongly augmented images.

\subsection{Training Process}
The final objective function is formed by combining Eq.~\ref{eq:loss-clean}, Eq.~\ref{eq:loss-noisy}, and Eq.~\ref{eq:penalty},
\begin{equation} \label{eq:loss-final}
    L=L_c+L_n+\lambda \sum_{(x_i,y_i)\in \mathbb C} [v(x_i) + v(\hat x_i)].
\end{equation}
The training process of our method is composed of two stages. First, we warm up the network using the objective function in Eq.~\ref{eq:loss-penalized}. Then, we adopt the criterion in Eq.~\ref{eq:loss-match} to select out clean samples and confident noisy samples, and optimize the network according to Eq.~\ref{eq:loss-final}. Stochastic gradient descent (SGD) is applied for updating network parameters. The detailed training process is described in Alg.~\ref{alg:train}.

\begin{algorithm}[t]
\caption{Algorithm for training the classification model via our proposed method. $E_1$ and $E_2$ denote the number of training epochs in the first and second stage respectively. $B$ means the number of batches.  }\label{alg:train}
\begin{algorithmic}[1]
\REQUIRE Training dataset $\mathbb D=\left\{(\mathbf x_i\textrm{,}\; y_i)\right\}_{i=1}^N$.
\ENSURE Optimized network parameters $\theta$.

\STATE Initialize $\mathbf q$ and  $\hat{\mathbf q}$: 
$q[k] = \hat q[k] = \frac{N_k}{N}$;

\FOR{$e\gets1$ to $E_1$}
\FOR{$b\gets1$ to $B$}

\STATE Fetch a batch of samples $\{(x_i, y_i)\}_{i=1}^n$;

\STATE Forward each sample $x_i$ through the network:\\
$\mathbf p_i=h(x_i\mid\theta)$;

\STATE Calculate the training loss with Eq.~\ref{eq:loss-penalized}:\\
$L=\frac{1}{n}\sum_{i=1}^n \ell_p(x_i,y_i)$;

\STATE Use SGD to update network parameters $\theta$;

\STATE Update the prior category distribution:\\ $\mathbf q \gets (1-\eta) \mathbf q + \eta \sum_{i=1}^n \mathbf p_i / n$;   
\ENDFOR
\ENDFOR

\FOR{$e\gets1$ to $E_2$}
\FOR{$b\gets1$ to $B$}
\STATE Fetch a batch of samples $\{(\mathbf x_i, y_i)\}_{i=1}^n$;
\STATE Augment all samples, resulting in $\{\hat{\mathbf x}_i\}_{i=1}^n$;
\STATE Forward each sample $x_i$ and $\hat x_i$: \\$\mathbf p_i=h(x_i\mid\theta)$, $\hat{\mathbf p}_i=h(\hat x_i\mid\theta)$;
\STATE Calculate loss value for each sample via Eq.~\ref{eq:loss-match}: \\\quad $\ell(x_i,y_i)=\ell^c(x_i,y_i)+\ell^c(\hat x_i, y_i)+\ell^c(\hat x_i, y_i^\prime)$;
\STATE Select out the set of clean samples:\\ $\mathbb C=\{(x_i,y_i)\mid\ell(x_i,y_i)<\epsilon\}$ \\where $\epsilon$ is the OTSU threshold;
\STATE Select out the set of confident noisy samples $\mathbb N$,  according to Eq.~\ref{eq:noisy};
\STATE Calculate overall training loss via Eq.~\ref{eq:loss-final};
\STATE Use SGD to update network parameters $\theta$;
\STATE Update the prior category distribution:\\
\STATE $\mathbf q \gets (1-\eta) \mathbf q + \eta \sum_{(x_i,y_i) \in \mathbb C} h(x_i\mid\theta) / \mid\mathbb C\mid$,\\ $\hat{\mathbf q} \gets (1-\eta) \hat{\mathbf q} + \eta \sum_{(x_i,y_i) \in \mathbb C} h(\hat x_i\mid\theta) / \mid\mathbb C\mid$; 
\ENDFOR
\ENDFOR
\RETURN Network parameters $\theta$.
\end{algorithmic}
\end{algorithm}
\section{Experiments}
\subsection{Datasets} 
Extensive experiments are conducted on 4 datasets, including CIFAR-10~\cite{2009Learning}, CIFAR-100, MNIST~\cite{1998Gradient}, and FashionMNIST~\cite{xiao2015learning} with synthetic noisy labels and distribution imbalance, and one real-word dataset Clothing1M~\cite{xiao2015learning}, to validate the effectiveness of our method. 

\begin{itemize}
    \item \textbf{CIFAR-10} is composed of 50,000 and 10,000 32$\times$32 images for training and testing respectively. The number of classes is 10.
    \item \textbf{CIFAR-100} is also composed of 50,000 and 10,000 32$\times$32 images for training and testing respectively. The number of classes is 100.
    \item \textbf{MNIST} is a digit recognition dataset consisting of 60,000 and 10,000 28$\times$28 grayscale images for training and testing respectively.
    \item \textbf{FashionMNIST} contains 60,000 training and 10,000 testing grayscale images with the size of 28$\times$28, which belong to 10 types of clothing. 
    \item \textbf{Clothing1M} contains 1,000,000 images with real-word label noises, which are regarded as the training set. 10,000 images with clean labels are used for testing.
\end{itemize}

First, to synthesize distribution imbalance in CIFAR-10, CIFAR-100, MNIST, and FashionMNIST, we randomly re-sample training images of all classes to make them obey a long-tailed distribution. The imbalance ratio is denoted as $\rho=\frac{\min_k N_k}{\max_k N_k}$.  
Following~\cite{han2018co}, we adopt two types of label corruption strategies to synthesize label noises: 1) Class-independent label noises are generated via randomly changing the label of every sample with a probability of $\eta$ (denoted as the ratio of noisy labels). If one label is determined to be modified, we replace it with a new label randomly drawn from other labels. 2) Class-dependent label noises are created through modifying the label of samples in the $k$-th class to $k+1$ with the probability of $\eta$. The classification accuracy is adopted for evaluation.

\subsection{Implementation Details}
Our method is implemented under PyTorch. During training, the weak augmentation is formed by random cropping and flipping, while AutoAugment~\cite{cubuk2018autoaugment} is utilized to generate strongly augmented images. We adopt ResNet32~\cite{he2016deep} as the backbone of the classification network. The learning rate of the SGD optimizer is initialized as 0.05 and decayed in the cosine annealing schedule~\cite{loshchilov2016sgdr}.
The training epochs in two stages are both 100, and the mini-batch size is 128. Without specification, all models are trained from scratch. Besides, we empirically set $\alpha$=2, $\kappa$=0.8, $\tau$=0.5, and $\lambda$=0.1.
8 V100 GPU-s are used to train models on Clothing1M. For other datasets, one GeForce RTX 2080Ti is used.
The training process costs around two days on Clothing1M, and around two hours on other datasets.

\begin{table}[t]
\caption{Comparison with existing LLN methods on CIFAR-10 with class-independent label noises ($\eta$=20\%). 
}
\centering
\setlength\tabcolsep{2pt}
\label{tab:lln-ci}
\begin{tabular}  {l|c|c|c|c|c}
\toprule 
\centering
Imbalance Factor $\rho$ &  1 & 0.1 & 0.05 & 0.02 & 0.01 \\ \midrule 
\textbf{CE} & 86.35 & 75.88 & 70.32 & 64.69 & 54.07 \\ 
\textbf{DivideMix}~\cite{li2020dividemix} & 84.03 & 69.80 & 60.26 & 54.74 & 36.56 \\
\textbf{Co-Teaching}~\cite{han2018co} & 89.75 & 58.63 & 52.58 & 42.67 & 39.39\\ 
\textbf{Co-Teaching+}~\cite{yu2019does}  & 88.36 & 42.58 & 35.26 & 33.90 & 34.88\\ 

\textbf{Co-Matching}~\cite{lu2021co} & 91.21 & 76.14 & 65.33 & 61.83 & 54.05 \\ 
\textbf{Co-Matching+OPP} & 90.57 & 81.86 & 77.58 & 70.60 & 57.71  \\ 

\textbf{SOA}~\cite{zhu2021second} & 90.36 & 79.02 & 67.56 & 58.44 & 51.70 \\

\textbf{JNPL}~\cite{kim2021joint} & 91.26 & 75.38 & 66.87 & 51.01 & 44.23 \\


\textbf{REL}~\cite{xia2020robust} & 84.71 & 69.00 & 65.71 & 46.73 & 36.31 \\
\textbf{DDLS}~\cite{zhang2021delving} & 89.88 & 74.59 & 63.45 & 43.33 & 35.14  \\
\textbf{AutoDo}~\cite{gudovskiy2021autodo} & 89.50 & 80.35 & 75.67 & 69.09 & 62.29\\
\textbf{LFDLN}~\cite{zhang2021learning} & 89.03 & 80.21 & 62.85 & 54.02 & 43.23 \\
\textbf{JoCoR}~\cite{wei2020combating}  & 90.21 & 81.63 & 74.15 & 63.87 & 53.69\\ 
\textbf{JoCoR+OPP} & 90.71 & 82.52 & 77.60 & 65.37 & 58.14 \\ 
\textbf{MOIT}~\cite{ortego2021multi} & 86.27 & 81.59 & 75.16 & 66.40 & 59.18  \\
\textbf{Jo-SRC}~\cite{yao2021jo} & 91.01 & 80.43 & 75.21 & 66.11 & 55.97 \\
\textbf{AUGDESC}~\cite{nishi2021augmentation} & 91.92 & 86.06 & 55.47 & 35.07 & 30.37 \\
\midrule
\textbf{Ours} &  \textbf{92.32} & \textbf{86.12} & \textbf{84.33} & \textbf{81.42} & \textbf{76.94} \\ \bottomrule
\end{tabular}
\end{table}
\begin{table}[h]
\caption{Comparison with existing LLN methods on CIFAR-10 with class-dependent noisy labels ($\eta$=20\%). 
}
\centering
\setlength\tabcolsep{2pt}
\label{tab:lln-cd}
\begin{tabular}  {l|c|c|c|c|c}
\toprule 
\centering
Imbalance Factor $\rho$ &  1 & 0.1 & 0.05 & 0.02 & 0.01 \\ \midrule 
\textbf{CE}       & 87.55 & 78.82 & 73.79 & 66.10 & 60.41 \\ 
\textbf{DivideMix}~\cite{li2020dividemix} &85.32 & 68.97 & 58.45 & 52.70 & 37.24 \\
\textbf{Co-Teaching}~\cite{han2018co}     & 89.14 & 46.34 & 41.43 & 36.65 & 34.99 \\ 
\textbf{Co-Teaching+}~\cite{yu2019does}     & 80.92 & 45.95 & 34.74 & 32.45 & 27.74 \\ 
\textbf{JoCoR}~\cite{wei2020combating}     & 89.68 & 57.43 & 49.01 & 41.98 & 41.41 \\ 
\textbf{JoCoR+OPP}    & 89.91 & 69.80 & 60.39 & 45.10 & 41.09 \\ 
\textbf{Co-Matching}~\cite{lu2021co}     & 82.47 & 58.86 & 50.40 & 43.31 & 35.95 \\
\textbf{Co-Matching+OPP}   & 89.48 & 70.86 & 60.69 & 47.05 & 40.82 \\ 
\textbf{AUGDESC}~\cite{nishi2021augmentation} & 91.54 & 60.52 & 39.61 & 34.11 & 29.93 \\
\textbf{SOA}~\cite{zhu2021second} & 87.87 & 78.94 & 64.32 & 60.58 & 49.24 \\

\textbf{JNPL}~\cite{kim2021joint} & 91.32 & 72.71 & 59.97 & 46.22 & 45.03 \\

\textbf{LFDLN}~\cite{zhang2021learning} & 83.38 & 76.36 & 56.22 & 49.31 & 48.68 \\
\textbf{REL}~\cite{xia2020robust} & 82.59 & 70.47 & 65.54 & 48.99 & 37.47 \\
\textbf{DDLS}~\cite{zhang2021delving} & 89.80 & 78.41 & 59.02 & 43.82 & 40.77 \\
\textbf{Jo-SRC}~\cite{yao2021jo} & 91.01 & 80.43 & 75.21 & 66.11 & 55.97 \\

\textbf{AutoDo}~\cite{gudovskiy2021autodo} & 90.71 & 81.33 & 78.10 & 71.85 & 69.09\\
\textbf{MOIT}~\cite{ortego2021multi} & 86.93 & 81.19 & 77.26 & 73.76 & 63.87 \\
\midrule
\textbf{Ours}       & \textbf{92.05} & \textbf{86.13} & \textbf{82.86} & \textbf{79.44} & \textbf{73.32}  \\ \bottomrule
\end{tabular}
\end{table}
\begin{figure*}[t]
\centering
\includegraphics[clip, trim=0.3cm 0.4cm 0.3cm 6.2cm, width=\textwidth]{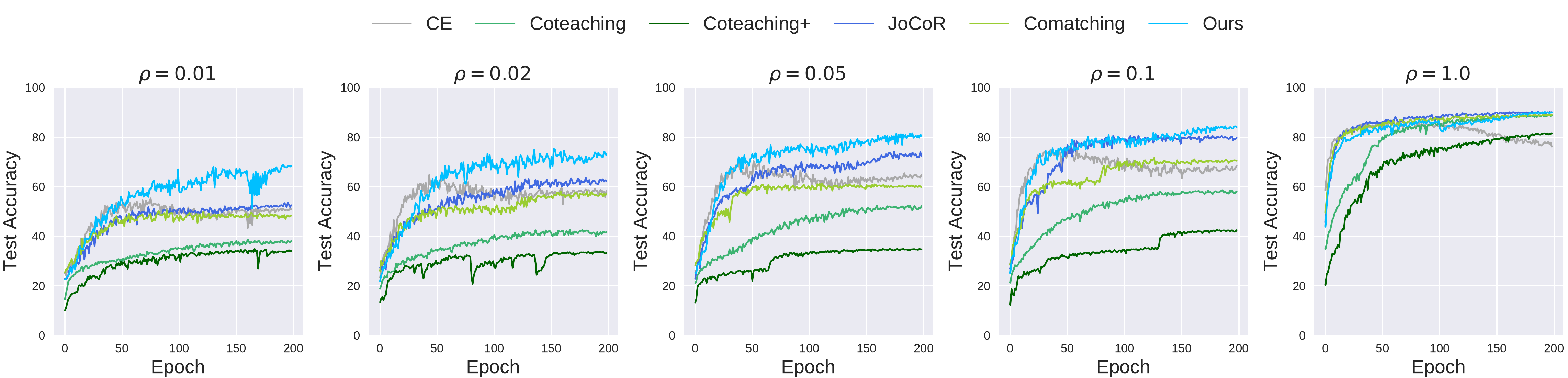}
\caption{Accuracy curves of methods for learning with class-independent label noise on CIFAR-10 with different imbalance ratios ($\eta = 20\%$).} 
\label{fig:imb}  
\end{figure*}

\begin{figure*}[t]
\centering
\begin{minipage}[c]{0.02\textwidth}
\begin{turn}{90} 
\textbf{CI Label Noise}
\end{turn}
\end{minipage}
\begin{minipage}[c]{0.97\textwidth}
\begin{tabular}{ >{\centering}p{0.225\textwidth}>{\centering}p{0.225\textwidth}>{\centering}p{0.225\textwidth}>{\centering}p{0.225\textwidth}}
CIFAR-10  & CIFAR-100 & MNIST & FashionMNIST
\end{tabular}
\includegraphics[clip, trim=1.8cm 0.8cm 0.1cm 2cm, width=0.24\textwidth]{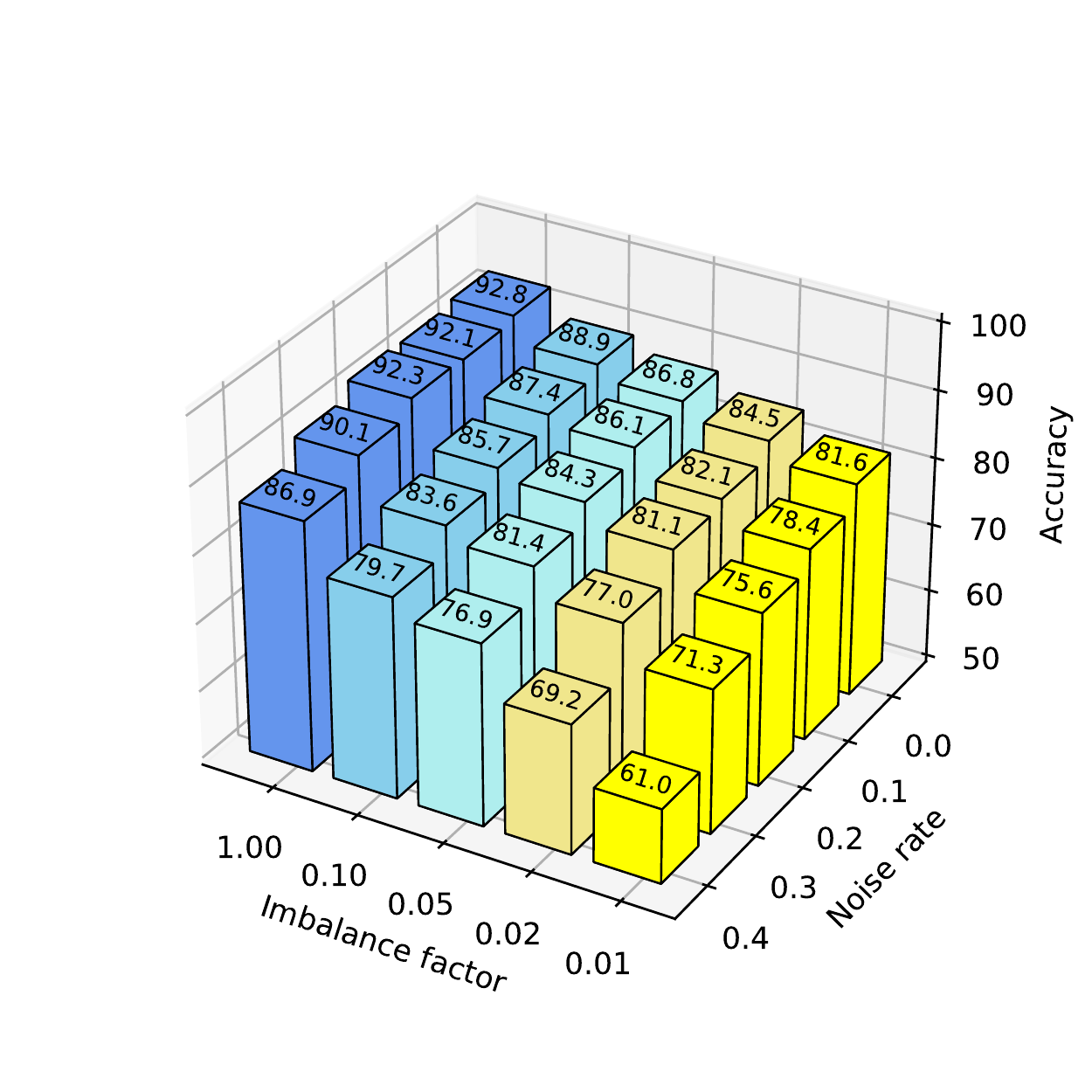}
\includegraphics[clip, trim=1.8cm 0.8cm 0.1cm 2cm, width=0.24\textwidth]{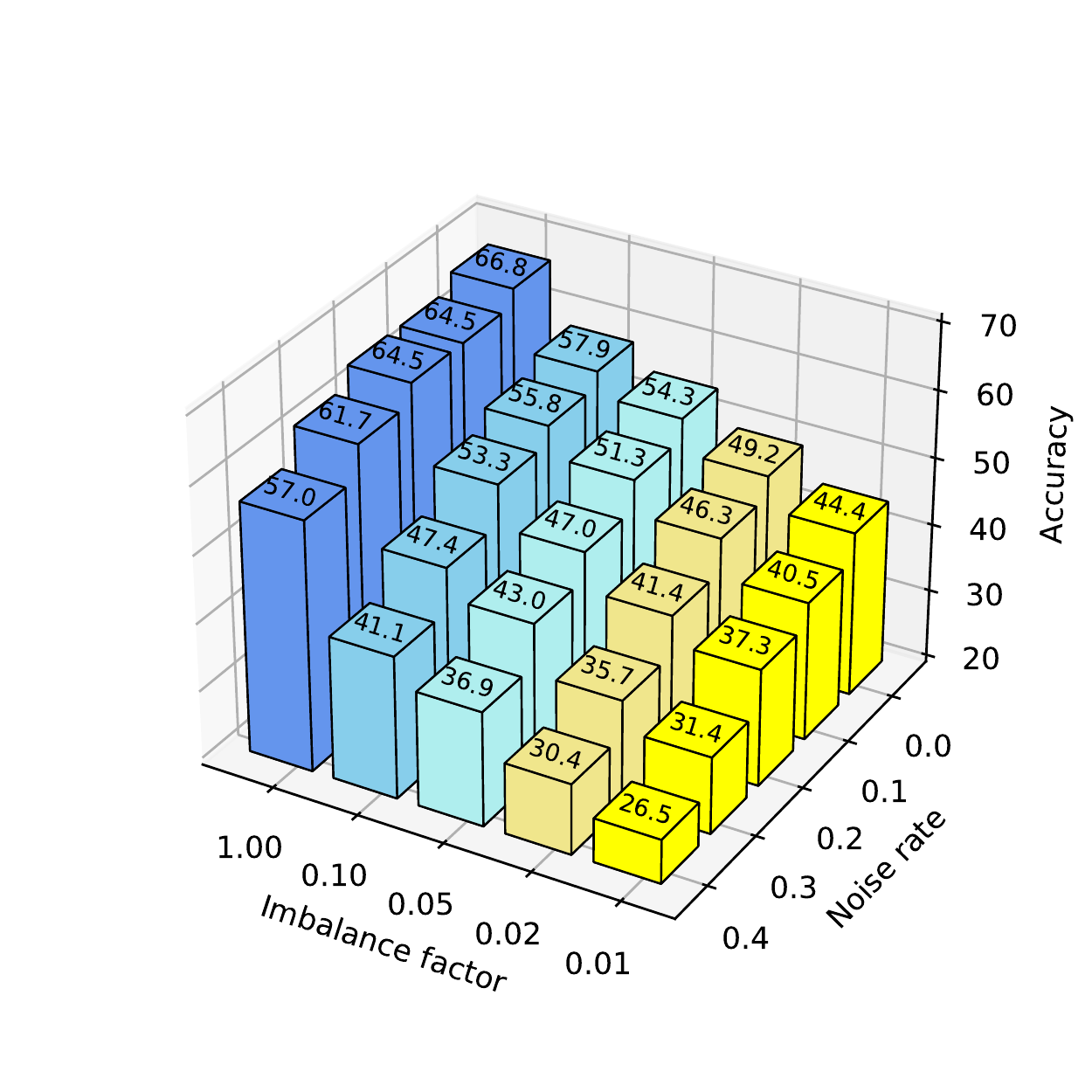}
\includegraphics[clip, trim=1.8cm 0.8cm 0.1cm 2cm, width=0.24\textwidth]{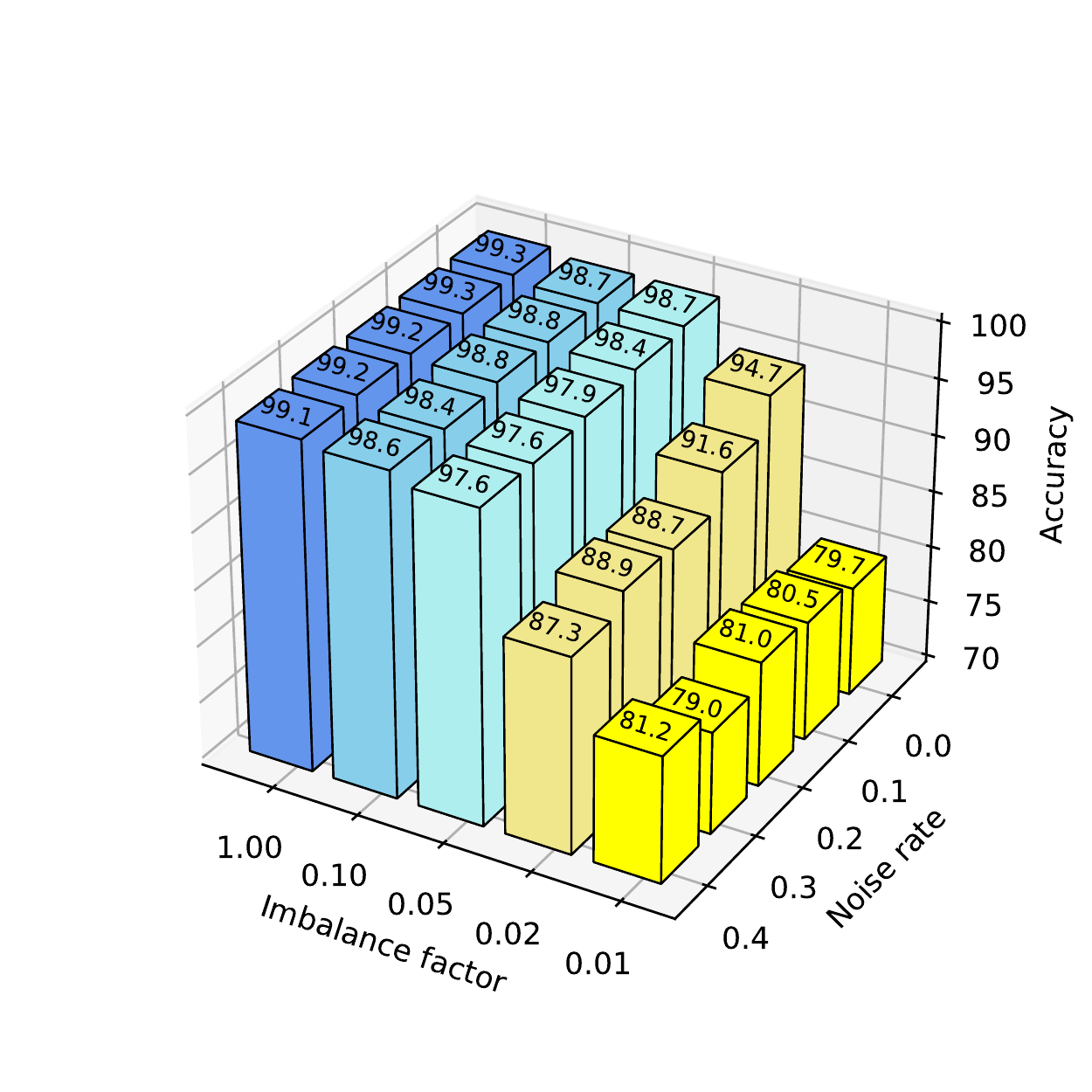}
\includegraphics[clip, trim=1.8cm 0.8cm 0.1cm 2cm, width=0.24\textwidth]{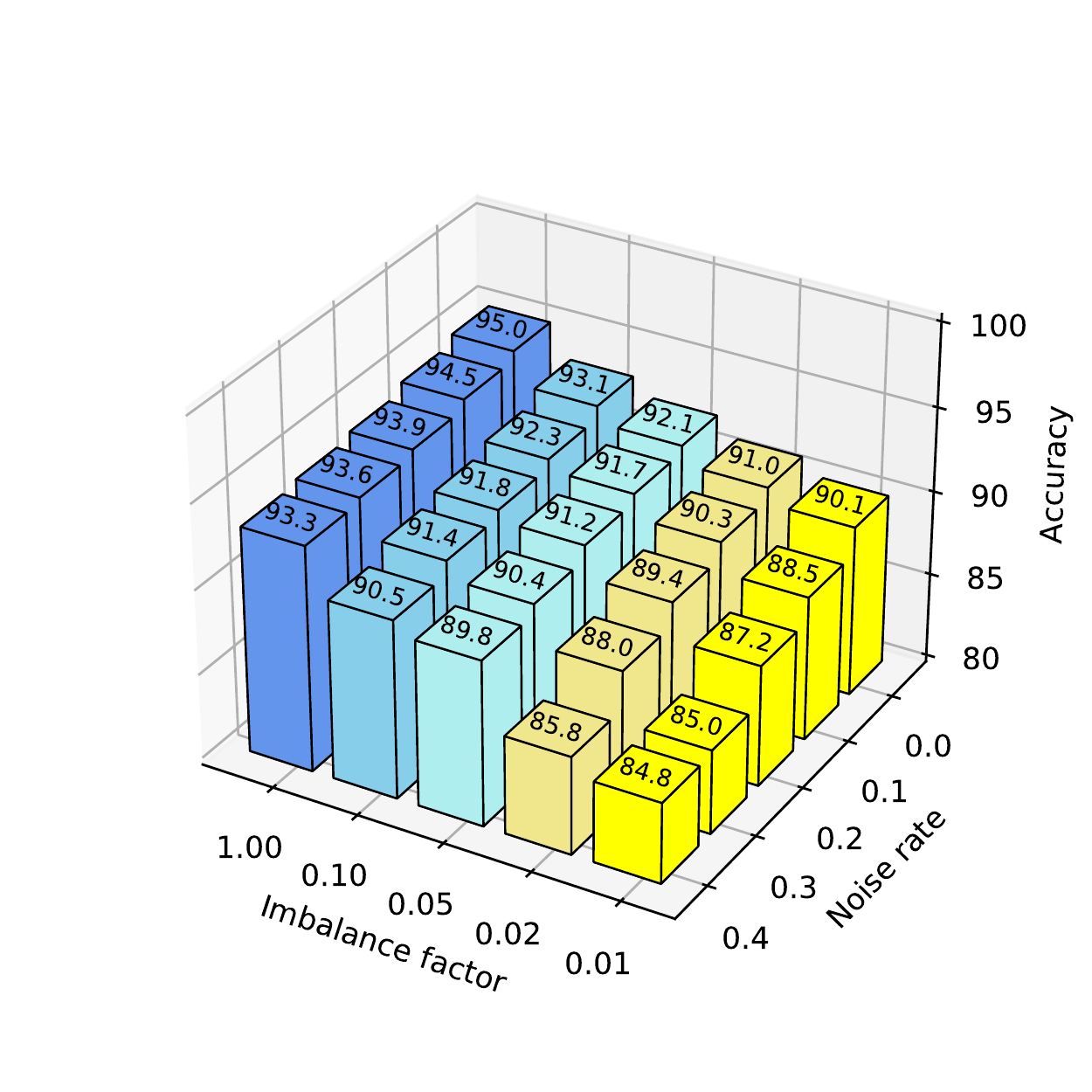}
\end{minipage}
\begin{minipage}[c]{0.02\textwidth}
\begin{turn}{90} 
\textbf{CD Label Noise}
\end{turn}
\end{minipage}
\begin{minipage}[c]{0.97\textwidth}
\includegraphics[clip, trim=1.8cm 0.8cm 0.1cm 2cm, width=0.24\textwidth]{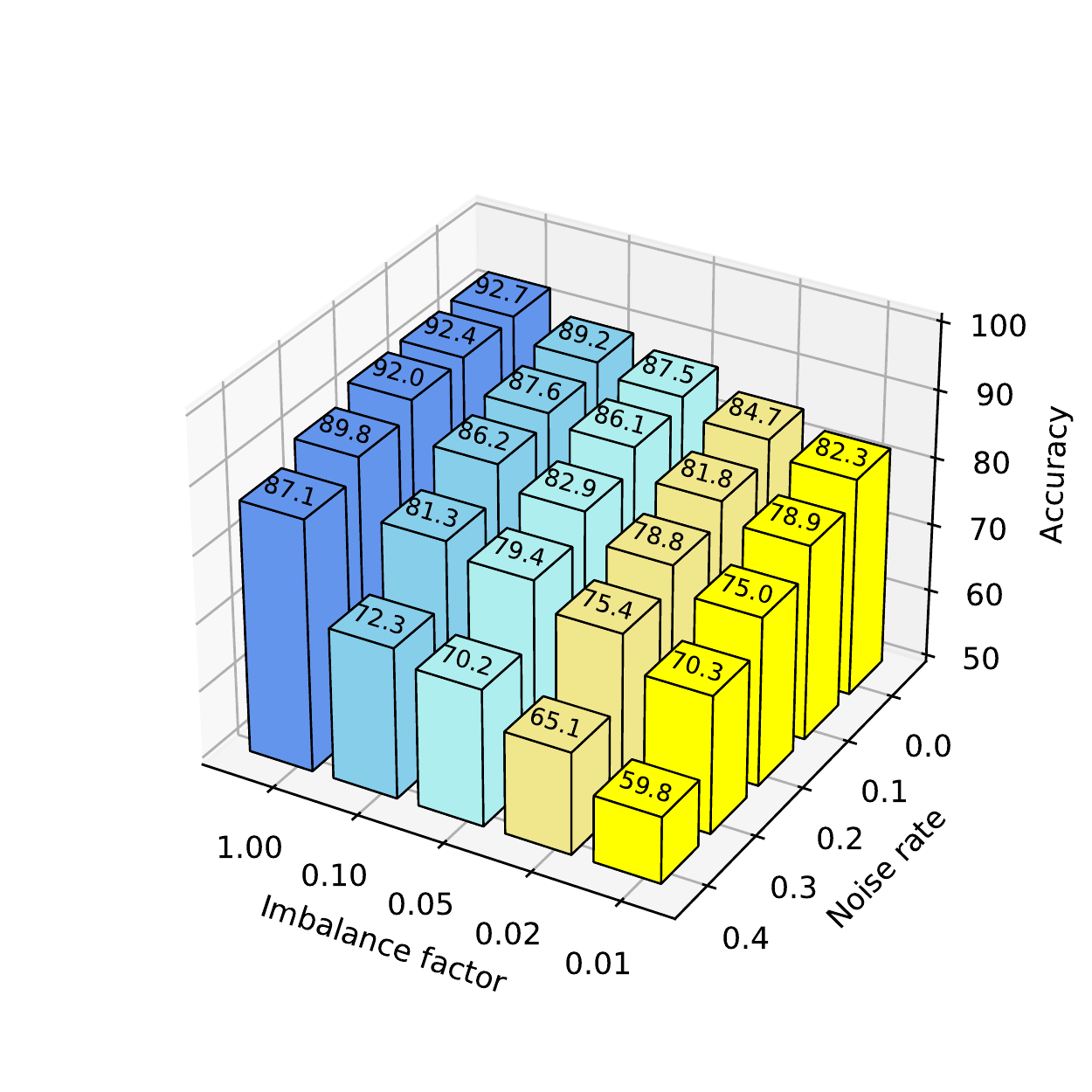}
\includegraphics[clip, trim=1.8cm 0.8cm 0.1cm 2cm, width=0.24\textwidth]{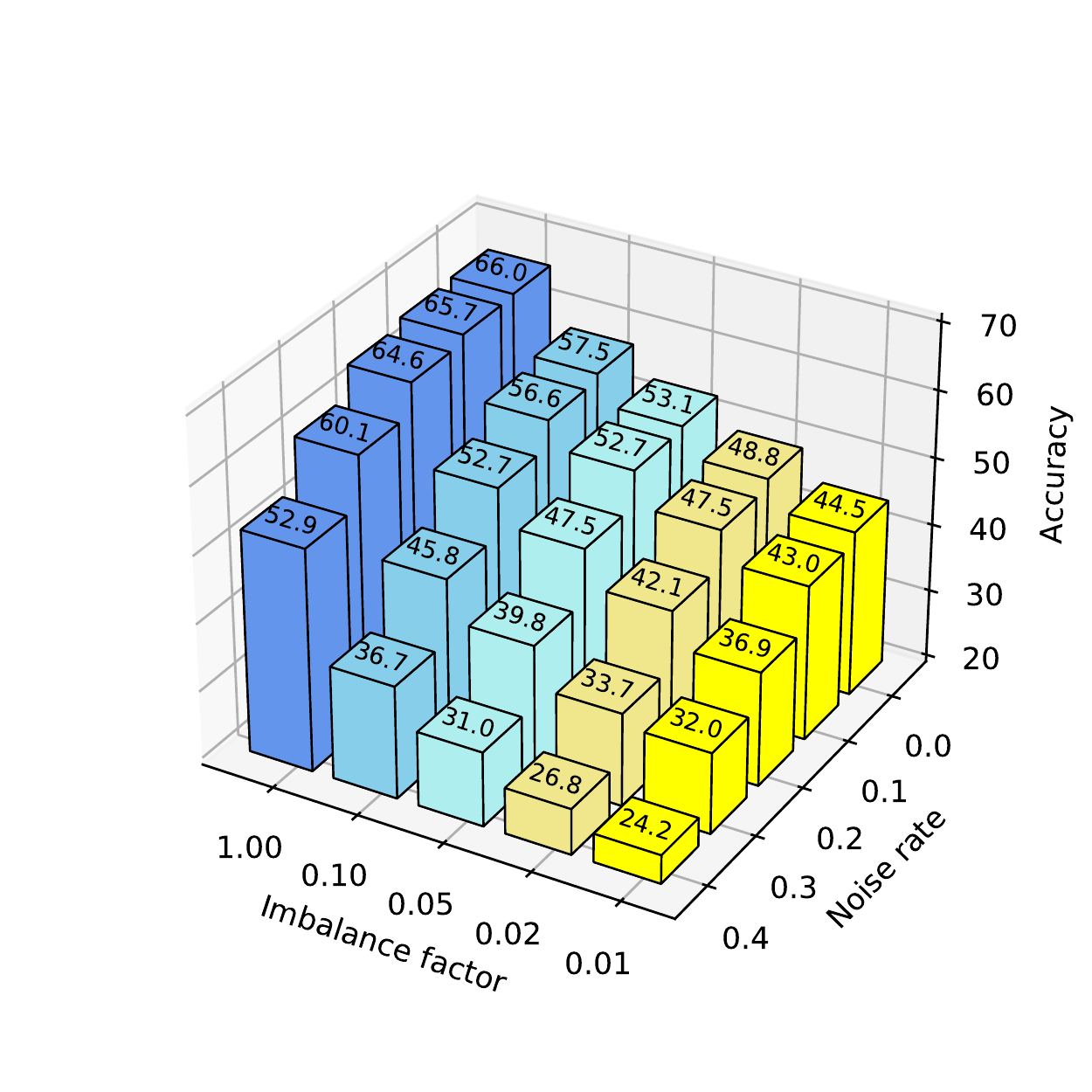}
\includegraphics[clip, trim=1.8cm 0.8cm 0.1cm 2cm, width=0.24\textwidth]{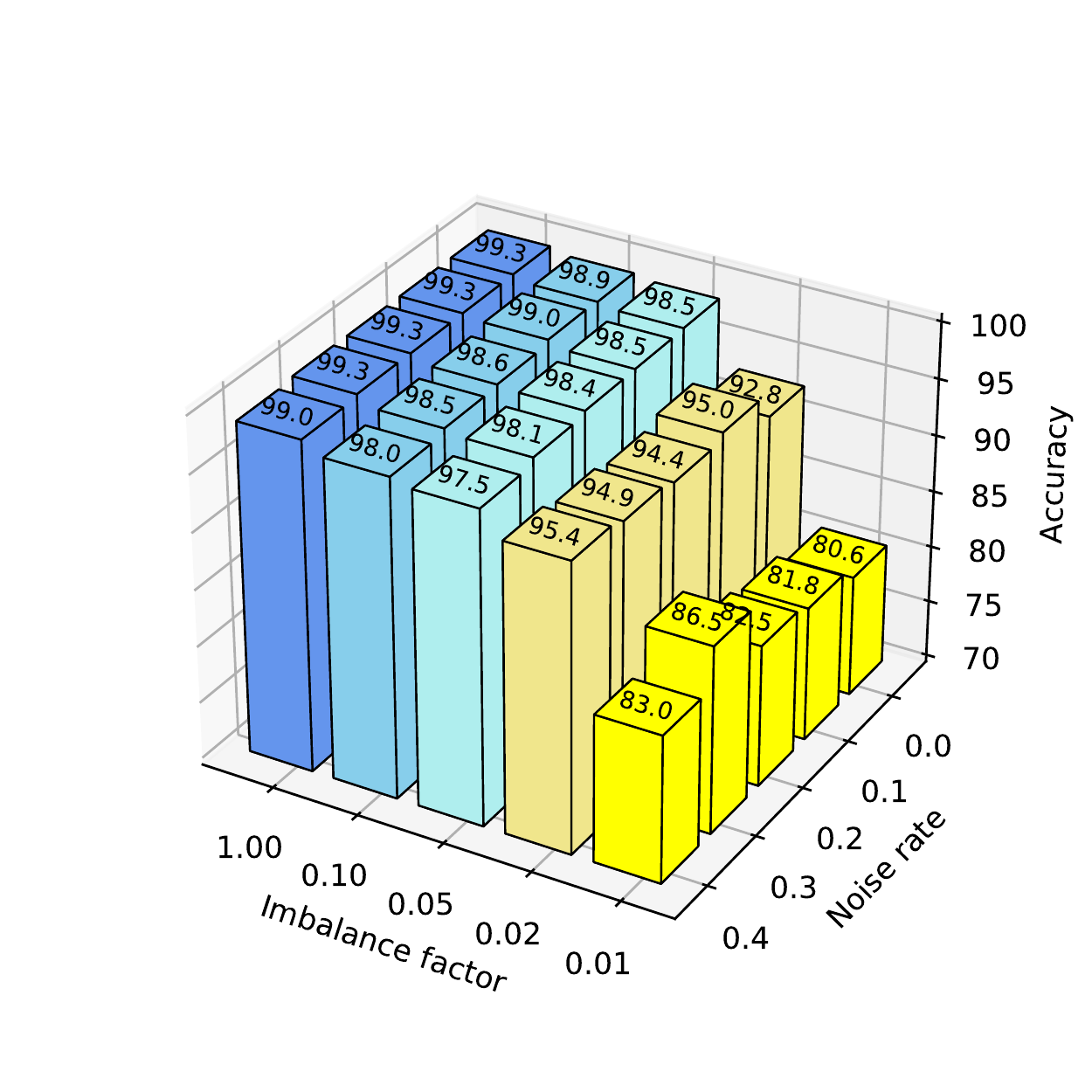}
\includegraphics[clip, trim=1.8cm 0.8cm 0.1cm 2cm, width=0.24\textwidth]{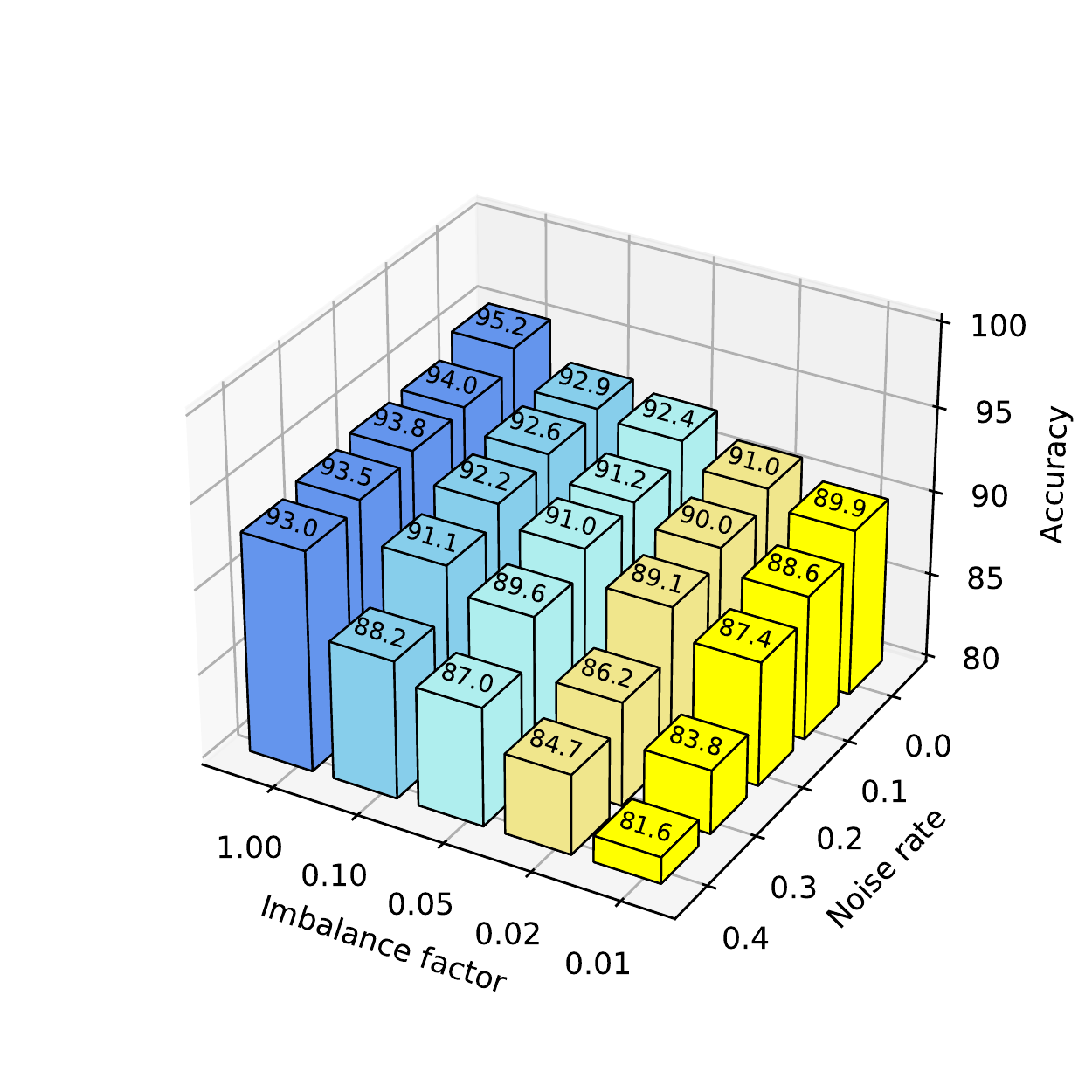}
\end{minipage}

\caption{Performance of our method on extensive settings for imbalance ratio and noise rate. `CI' and `CD' stands for `class-independent' and `class-dependent' respectively.
} 
\label{fig:imb-noise-2d}  
\end{figure*}
\subsection{Comparison against Methods for Learning with Label Noise}
\subsubsection{Existing Methods}
We compare our method against the following  methods for learning with label noise (LLN). 

\begin{itemize}
    \item[1)] \textbf{Joint Optimization}~\cite{tanaka2018joint} proposes to alternatively optimize network parameters and update labels according to current network predictions.
    \item[2)] \textbf{Co-Teaching}~\cite{han2018co} makes usage of the divergence in the learning errors of two distinct models to remove the influence of label noises. The two models screen out noisy samples for each other, combining their advantages in counteracting noisy labels. 
    \item[3)] \textbf{Co-Teaching+}~\cite{yu2019does} is an extended variant of \textbf{Co-Teaching}. It utilizes the `update by disagreement' strategy to prevent the co-teaching method degenerating into a self-training method, since the two models converge to a consensus in late training epochs.
    \item[4)] \textbf{MLNT}~\cite{li2019learning} devises a beforehand meta-learning algorithm to make the network parameters tolerant to label noises. In meta-train, noisy labels are used for updating network parameters. In meta-test, the mean teacher is used for correcting the optimization direction.
    \item[5)] \textbf{MetaCleaner}~\cite{zhang2019metacleaner} proposes to construct clean samples via aggregating noisy-labeled samples with estimated importance scores.
    \item[6)] \textbf{PENCIL}~\cite{yi2019probabilistic} leverages the probabilistic modeling to jointly optimize network parameters and infer the latent correct labels.
    \item[7)] \textbf{JoCoR}~\cite{wei2020combating} adopts two networks to identify clean samples via a joint loss which is composed of individual cross entropy losses and a co-regularized Jensen-Shannon divergence loss.
    \item[8)] \textbf{DivideMix}~\cite{li2020dividemix} separates clean and noisy samples via a two-component Gaussian mixture model built upon the training loss. Then, the MixMatch algorithm~\cite{berthelot2019mixmatch} is employed to explore noisy samples in a semi-supervised manner.  
    \item[9)]  \textbf{ELR+}~\cite{liu2020early} reveals that semi-supervised learning regularizations are capable of neutralizing the negative influence of noisy samples.
    \item[10)]  \textbf{Co-Matching}~\cite{lu2021co} utilizes the matching between inferences on weakly and strongly augmented images to discern noisy samples. Separate models are used to infer recognition results for weakly and strongly augmented images.
    \item[11)] \textbf{MOIT}~\cite{ortego2021multi} proposes an interpolated contrastive loss for alleviating the side effect of label noises in representation learning. A k-nearest neighbor search algorithm is devised for detecting noisy samples, which are subsequently explored via semi-supervised learning. 
    
    \item[12)]  \textbf{AUGDESC }~\cite{nishi2021augmentation} finds that choosing different augmentation policies to model the loss metric for noisy sample detection and learn the classifier is more effective in coping with noisy labels. It also reveals that applying strong augmentation in the warm-up training stage encourages the network to memorize noisy data soon.
    
    \item[13)] \textbf{SOA}~\cite{zhu2021second} 
    involves in second-order statistics (namely covariance terms) to deal with instance-dependent label noises.
    \item[14)]  \textbf{Jo-SRC}~\cite{yao2021jo} 
    leverages the Jensen-Shannon divergence between predictions and labels to recognize noisy samples. Besides, the prediction consistency between two views of each sample is adopted for discriminating in-distribution noisy samples from out-of-distribution noisy samples, and a mean teacher model is built for rectifying the labels of in-distribution samples. 
    
    \item[15)] \textbf{JNPL}~\cite{kim2021joint} introduces a joint negative and positive learning framework. For preventing the underfitting of training data, the method decreases weights of negative learning terms with respect to predictions on negative labels. Furthermore, it adopts a strict sampling strategy to select positive samples, and reweights their losses in positive learning according to the maximum predicted probabilities. 
    
    \item[16)] \textbf{AutoDo}~\cite{gudovskiy2021autodo} devises a generalized automated dataset optimization algorithm, aiming to eliminate the distribution gap between testing data and training data with noisy labels.
    \item[17)]  \textbf{LFDLN}~\cite{zhang2021learning} proposes to correct labels of samples progressively, starting from those samples with high prediction confidences. 

    \item[18)]  \textbf{REL}~\cite{xia2020robust} 
    separates network parameters into two groups, including critical and non-critical parameters. 
    The latter type of parameters are decayed in the early learning strategy for eliminating the influence of noisy samples.
    \item[19)] \textbf{DDLS}~\cite{zhang2021delving} presents an online label
    smoothing strategy, which assigns a soft probability distribution to target and non-target classes based on the accumulation of network predictions.
    
\end{itemize}

\subsubsection{Experiments on CIFAR-10}
The experimental results on CIFAR-10 with class-independent noisy labels and class-dependent noisy labels are presented in Table~\ref{tab:lln-ci} and~\ref{tab:lln-cd}, respectively. The noise rate $\eta$ is set to 20\%, and the imbalance ratio $\rho$ varies in $\{1,10,20,50,100\}$. \textbf{CE} indicates the baseline model trained with the standard cross-entropy loss function.
It can be evidently observed that, without specific designs for processing label noises, existing LLN methods are severely influenced by the data imbalance.
For instance, on CIFAR-10 with class-independent label noises, \textbf{JoCoR}, \textbf{Co-Matching} and \textbf{AUGDESC} achieve promising performance (90.21\%, 91.21\%, and 91.92\% respectively) when data is balanced ($\rho$=1). However, when data is drastically imbalanced ($\rho$=0.01), their accuracy values are drastically decreased by 37.16\%, 36.52\%, and 61.55\%, respectively.
When the imbalance ratio $\rho$ is smaller than 0.02, all of the existing methods even perform worse than \textbf{CE}. Our method is effective in coping with label noises, and derives of substantially better performance than other methods. For example, on CIFAR-10 with class-independent label noises, the accuracy of our method (86.12\%) is 10.45\% higher than that of the second-best method \textbf{AutoDo} (75.67\%) when $\rho=0.05$; our method generates classification result with $14.65\%$ higher accuracy than \textbf{AutoDo} when $\rho=0.01$. Fig.~\ref{fig:imb} illustrates the variation curves of the test accuracy along the training epochs, indicating that our method outperforms other methods steadily in late training epochs. We also attempt to incorporate the data re-balancing strategy, namely online prior penalization (OPP), into existing LLN methods \textbf{JoCoR} and \textbf{Co-Matching}, forming \textbf{JoCor+OPP} and \textbf{Co-Matching+OPP}. The adoption of OPP helps improve \textbf{JoCoR} and \textbf{Co-Matching}. Meanwhile,  our proposed cross-augmentation matching strategy is much more robust against label noises under long-tailed distribution.


\subsubsection{Experiments on Clothing1M}
\begin{table}[t]
\caption{Comparison with LLN methods on Clothing1M.} \label{tab:sota}
\centering
\setlength\tabcolsep{4pt}
\begin{tabular}  {l|c}
\toprule
Method & Accuracy \\ \midrule
\textbf{CE} &  69.21 \\ 
\textbf{M-correction}~\cite{arazo2019unsupervised}& 71.00  \\ 
\textbf{Joint Optimization}~\cite{tanaka2018joint}& 72.16  \\ 
\textbf{MetaCleaner}~\cite{zhang2019metacleaner} & 72.50  \\ 
\textbf{MLNT}~\cite{li2019learning} & 73.47  \\ 
\textbf{PENCIL}~\cite{yi2019probabilistic} & 73.49  \\
\textbf{DivideMix}~\cite{li2020dividemix} & 74.76 \\
\textbf{ELR+}~\cite{liu2020early} & 74.81  \\
\textbf{AUGDESC}~\cite{nishi2021augmentation} & 75.11 \\ \midrule
\textbf{Ours} & \textbf{75.49} \\ \bottomrule
\end{tabular}
\end{table}
Experimental results on Clothing1M which has real-world label noises are presented in Table~\ref{tab:sota}. For fair comparison, ResNet50 which is initialized with ImageNet pre-trained parameters are used to implement our method. 
Results of \textbf{M-correction}~\cite{arazo2019unsupervised}, 
\textbf{Joint Optimization}~\cite{tanaka2018joint},
\textbf{MetaCleaner}~\cite{zhang2019metacleaner},
\textbf{MLNT}~\cite{li2019learning},
\textbf{PENCIL}~\cite{yi2019probabilistic},
\textbf{DivideMix}~\cite{li2020dividemix},
\textbf{ELR+}~\cite{liu2020early},
and \textbf{AUGDESC}~\cite{nishi2021augmentation} reported in their original papers are used.
Our approach achieves the best performance, surpassing the second best method \textbf{AUGDESC} by 0.38\%.

Exhaustive experiments on four datasets (CIFAR-10, CIFAR-100, MNIST, and FashionMNIST) with various imbalance factors and noise ratios are presented in Fig.~\ref{fig:imb-noise-2d}.


\begin{table}[t]
\caption{Comparison with existing LLTD methods on CIFAR-10 ($\rho$=0.01). 
`RS', `RW' and `LS' stands for the resampling strategy in~\cite{ren2018learning}, the reweighting strategy in~\cite{chawla2002smote}, and label smoothing stragey in~\cite{szegedy2016rethinking}, respectively. 
} \label{tab:ltl}
\centering
\setlength\tabcolsep{3.5pt}
\begin{tabular}  {l|c|c|c|c|c}
\toprule 
Noise Rate $\eta$ & 0.0 & 0.1 & 0.2& 0.3 & 0.4 \\ \midrule
\textbf{CE} & 69.85 & 59.85 & 54.94 & 48.38 & 48.27\\ 
\textbf{Mixup} & 73.06 & 60.90 & 48.11 & 45.94 & 43.92 \\ 
\textbf{LDAM-DRW} &78.90 & 68.24 & 60.69 & 56.50 & 48.15 \\
\textbf{MiSLAS} & 81.91 & 71.32 &  66.23 & 63.11 &  51.34 \\ \midrule
\textbf{Ours-RS}  & 82.97 & 77.28 & 73.04 & 62.55 & 47.82  \\
\textbf{Ours-RW} & 77.48 & 76.01 & 72.41 & 66.02 & 51.25 \\ 
\textbf{Ours-LS} & 68.97 & 69.17 & 66.25 & 62.28 & 54.86 \\ 
\textbf{Ours} & \textbf{86.87} & \textbf{79.66} & \textbf{76.94} & \textbf{69.23} & \textbf{61.03} \\ \bottomrule
\end{tabular}
\vspace{-1em}
\end{table}
\begin{figure*}[t]
\centering
\includegraphics[clip, trim=0.3cm 0.4cm 0.3cm 6.2cm, width=\textwidth]{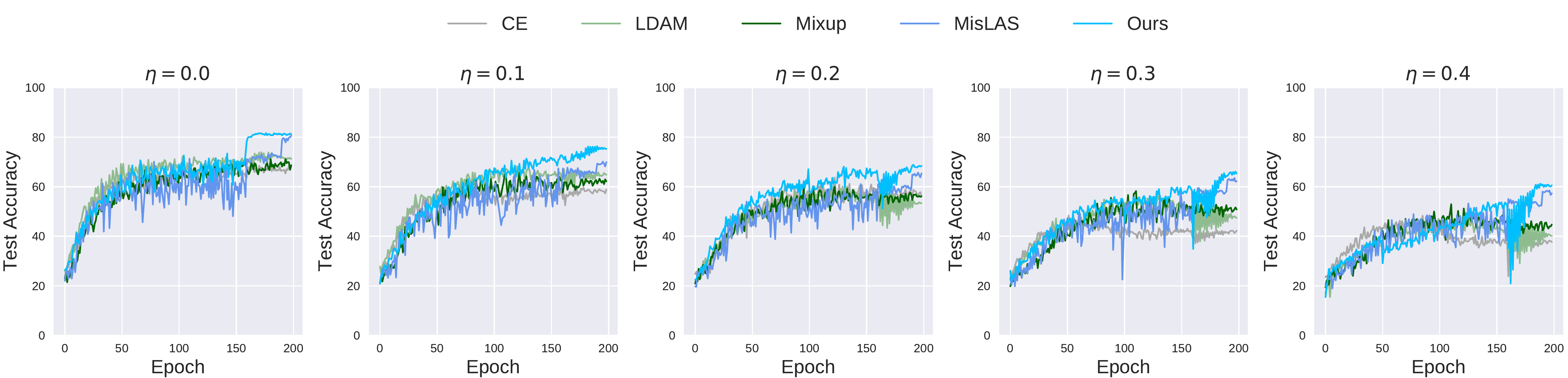}
\caption{Accuracy curves of methods for learning with long-tailed distribution on CIFAR-10 with $\rho=0.01$.} 
\label{fig:nr}  
\end{figure*}
\subsection{Comparison against Methods for Learning under Long-Tailed Distribution} 
In this subsection, we compare our method against existing methods capable of addressing the problem of learning under long-tailed distribution (LLTD), including \textbf{RW} (short for data reweighting)~\cite{chawla2002smote}, \textbf{LS}~\cite{szegedy2016rethinking}, \textbf{Mixup}~\cite{zhang2017mixup},  \textbf{RS} (short for data resampling)~\cite{ren2018learning}, \textbf{LDAM-DRW}~\cite{cao2019learning}, and \textbf{MiSLAS}~\cite{zhong2021improving}. 
\textbf{RW} can be easily implemented via assigning relatively higher weights to samples of tail classes. \textbf{RS} usually balances training data through under-sampling head classes or over-sampling tail classes. Other methods are briefly introduced as below.

\begin{itemize}
    \item[1)] \textbf{LS}~\cite{szegedy2016rethinking} (label smoothing) is a regularization technique that addresses both overfitting and overconfidence, which has been widely explored in the filed of image recognition.
    \item[2)] \textbf{Mixup}~\cite{zhang2017mixup} is a data augmentation technique that generates a weighted combinations of random image pairs from the training data. Given two images and their ground truth labels, ($\mathbf x_i, \mathbf y_i$) and ($\mathbf x_j, \mathbf y_j$), a synthetic training example ($\hat{\mathbf x},\hat{\mathbf y}$) is generated as:\\
    \begin{align}
    &\hat{\mathbf x} = \hat{\lambda} \mathbf x_i + (1 - \hat{\lambda}) \mathbf x_j \\
    &\hat{\mathbf y} = \hat{\lambda} \mathbf y_i + (1 - \hat{\lambda}) \mathbf y_j
    \end{align}
    $\hat{\lambda}$ is randomly drawn from a beta distribution.
    \item[3)] \textbf{LDAM-DRW}~\cite{cao2019learning} proposes a label-distribution-aware margin (LDAM) loss for improving the generalization to minority classes. This work reveals that deferring the rebalancing optimization helps to relieve the complications of re-balancing algorithms.
    \item[4)] \textbf{MiSLAS}~\cite{zhong2021improving} utilizes label-aware smoothing strategies to calibrate distributions of predicted probability values. Its training phase is also composed of two stages, including a first instance-balanced stage and a second class-balanced stage. It fixes the transformation parameters and merely adjusts the mean and variance when updating  the batch normalization modules in the second stage.
\end{itemize}

The results are presented in Table~\ref{tab:ltl}, and the variation curves of test accuracy are illustrated in Fig.~\ref{fig:nr}. Our method has better generalization ability than state-of-the-art LLTD approaches. We also attempt to replace the online prior penalization with \textbf{RS}~\cite{ren2018learning}, \textbf{RW}~\cite{chawla2002smote}, or \textbf{LS}~\cite{szegedy2016rethinking} for re-balancing the clean sample set $\mathbb C$, producing results indicated by \textbf{Our-RS}, \textbf{Our-RW}, or \textbf{Our-LS}, respectively. These data rebalancing strategies are inferior to the online prior penalization. 

\begin{table}[t]
\caption{Comparison of our method against naïve combinations of methods for LLN and LLTD separately, on CIFAR-10 with class-independent label noises ($\rho=100$ and $\eta=0.2$).  }\label{tab:naive}
\vspace{-0.5em}
\centering
\setlength\tabcolsep{6pt}
\begin{tabular}  {l|c|c|c|c}
\toprule 
 (1) + (2) & RW & RS & LS & LDAM  \\ \midrule
\textbf{Co-Teaching} & 39.31 & 47.79 & 37.01 & 42.50 \\ 
\textbf{Co-Teaching+}  & 27.33 & 43.06 & 34.61 & 47.51 \\ 
\textbf{JoCoR}  & 57.20 & 60.04 & 23.46 & 58.40 \\ 
\textbf{Co-Matching} & 60.54 & 64.44 & 12.34 & 59.47  \\ 
\textbf{Ours} &  \textbf{72.41} & \textbf{73.04} & \textbf{66.25} & \textbf{76.94} \\
\bottomrule
\end{tabular}
\end{table}

\subsection{Comparison to Naïve Combinations of Existing LLN and LLTD Methods} 
We attempt to simply combine methods for learning with label noise including \textbf{Co-Teaching}~\cite{han2018co}, \textbf{Co-teaching+}~\cite{yu2019does}, \textbf{JoCoR}~\cite{wei2020combating}, \textbf{Co-Matching}~\cite{lu2021co}, and methods for learning with long-tailed distribution including \textbf{RS}~\cite{ren2018learning}, \textbf{RW}~\cite{chawla2002smote}, \textbf{LS}~\cite{szegedy2016rethinking}, and \textbf{LDAM}~\cite{cao2019learning}. 
We also implement variants of our method via combining the cross-augmentation matching strategy with the above four methods for learning under long-tailed distribution.
Experimental results on CIFAR-10 with $\rho=0.01$ and $\eta=0.2$ are reported in Table~\ref{tab:naive}, and variants of our method outperform other methods consistently. For example, when incorporated with \textbf{LDAM}, our method achieves 34.44\%, 29.43\%, 18.54\%, and 17.47\% higher accuracy than \textbf{Co-Teaching}, \textbf{Co-Teaching+}, \textbf{JoCoR}, and \textbf{Co-Matching}, respectively.

 
\begin{table*}[t]
\caption{Comparison of variants of our method using different objective functions
. Four datasets are used in experiments.  $\eta=20\%$ and $\rho=0.01$. `CI' and `CD' indicate class-independent and class-dependent label noises are used respectively. `CE' indicates the baseline method trained with the cross-entropy loss calculated between predictions and given labels. 
}
\centering
\label{tab:ablation}
\setlength\tabcolsep{2.5pt}
\begin{tabular}  {l|c|c|c|c|c|c|c|c}
\toprule
\multirow{2}{*}{Method} & \multicolumn{2}{c|}{CIFAR-10} & \multicolumn{2}{c|}{CIFAR-100} & \multicolumn{2}{c|}{MNIST} & \multicolumn{2}{c}{FashionMNIST}\\

\cmidrule(l){2-3} \cmidrule(l){4-5} \cmidrule(l){6-7} \cmidrule(l){8-9}
 & CI & CD & CI & CD & CI & CD & CI & CD \\ \midrule



\textbf{CE}  & 54.07 & 62.57 & 24.64 & 26.01 & 95.83 & 96.77 & 82.36 & 84.40 \\ 
\textbf{CE} + \textbf{CAugMatching} & 61.47 & 65.85 & 30.84 & 31.88  & 96.79 & 97.33 & 85.15 & 85.89 \\ 
\textbf{CE} + \textbf{CAugMatching} + \textbf{LNOR} &  65.02 & 65.97  & 32.16 & 32.80  & 97.31 & 97.50  & 85.19 & 86.44  \\ 
\textbf{CE} + \textbf{CAugMatching} + \textbf{LNOR} + \textbf{OPP} & \textbf{72.86} & \textbf{72.83} & \textbf{34.89} & \textbf{34.67} & \textbf{97.92} & \textbf{97.77} & \textbf{86.35} & \textbf{86.72} \\ \bottomrule
\end{tabular}
\end{table*}

\subsection{Ablation Study}\label{sec:as}
We conduct elaborate ablation study to validate core components of our method. Without specification, the class independent label noise is used, and $\eta$ and  $\rho$ is set to $20\%$ and 0.01 respectively.

\subsubsection{Effectiveness of Loss Components}
\begin{figure}[t]
\centering
\includegraphics[clip, trim=0.3cm 0.6cm 0.2cm 0.5cm, width=0.492\columnwidth]{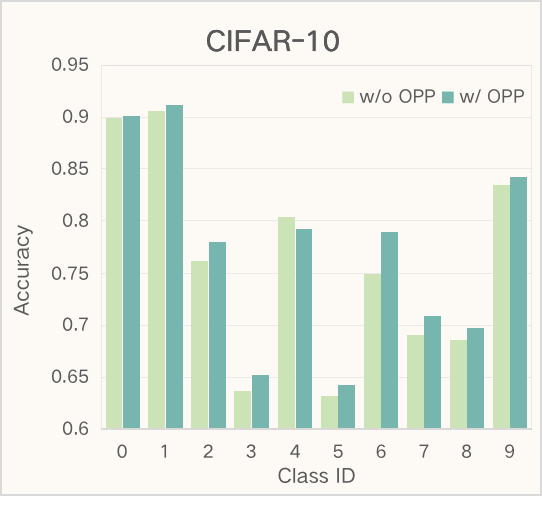}
\includegraphics[clip, trim=0.3cm 0.6cm 0.2cm 0.5cm, width=0.492\columnwidth]{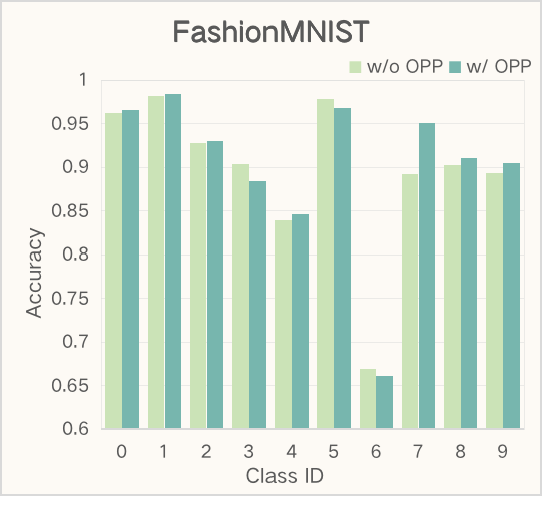}
\caption{Class-wise accuracy of using or not using online prior penalization (OPP) on CIFAR-10 and FashionMNIST datasets. $\rho=0.01$ and $\eta=20\%$. The class sample size decreases as the class ID grows. } 
\label{fig:opp}  
\end{figure}

Thorough ablation study is conducted to validate loss components in our method, including cross-augmentation matching (CAugMatching) in Eq.~\ref{eq:loss-match}, leave-noise-out regularization (LNOR) in Eq.~\ref{eq:loss-noisy}, and online prior penalization (OPP) in Eq.~\ref{eq:penalty}. The experimental result is presented in Table~\ref{tab:ablation}.
CE is used as the loss criterion for measuring the consistency between predictions and given labels.
As shown in Table~\ref{tab:ablation}, the adoption of CAugMatching brings significant performance improvement, because it is advantageous in discriminating noisy samples from clean samples and enhancing the representation capacity. 
For example, CAugMatching leads to an accuracy gain of 6.20\% on CIFAR-100 dataset with class-independent label noises.
The LNOR helps to mitigate the influence caused by noisy samples. It can also benefit the training of our model in all cases, e.g., the accuracy is improved by 3.55\% on CIFAR-10 with class-independent label noises.
The OPP loss, devised for balancing training data, can bring substantial accuracy gains. For example, on CIFAR-10 and FashinMNIST with class-independent label noises, the accuracy is increased by 7.84\% and 1.16\%, respectively. As can be observed from the class-wise accuracy in Fig.~\ref{fig:opp}, OPP is beneficial for improving the classification performance on tail classes.

\begin{table}[t]
\caption{Efficacy of our method applied with different loss functions. `Ours+CE', `Ours+Focal', and `Ours+LDAM' denotes variants of our method trained with cross entropy loss, focal loss, and LDAM loss, respectively.   
}\label{tab:criterion}
\centering
\begin{tabular}{ >{\centering}p{0.3\linewidth}>{\centering}p{0.25\linewidth}>{\centering\arraybackslash}p{0.25\linewidth}}
\toprule
\centering
Method               & CIFAR-10 & CIFAR-100 \\ \midrule
\textbf{CE}           & 54.07 & 24.64 \\ 
\textbf{Ours + CE}    & \textbf{72.86} & \textbf{34.89}  \\ \midrule
\textbf{Focal}        & 56.04 & 25.07  \\ 
\textbf{Ours + Focal} & \textbf{73.34} & \textbf{35.06}  \\ \midrule
\textbf{LDAM}         & 62.28 & 27.73  \\ 
\textbf{Ours + LDAM}  & \textbf{76.94} & \textbf{36.92} \\ \bottomrule
\end{tabular}
\end{table}


\subsubsection{Robustness to Loss Criteria}
We attempt to replace the cross entropy loss in Eq.~\ref{eq:loss-ce} with the focal loss~\cite{lin2017focal} or the LDAM loss. As shown in Table~\ref{tab:criterion}, our method is pretty stable when applied with different loss functions. For example, on the CIFAR-100 dataset, our method leads to the accuracy gain of 8.99\% and 6.80\%, when incorporated with the focal loss and the LDAM loss respectively. 


\begin{figure*}[t]
\centering
\includegraphics[clip, trim=0.0cm 0.3cm 0.0cm 0.0cm, width=\textwidth]{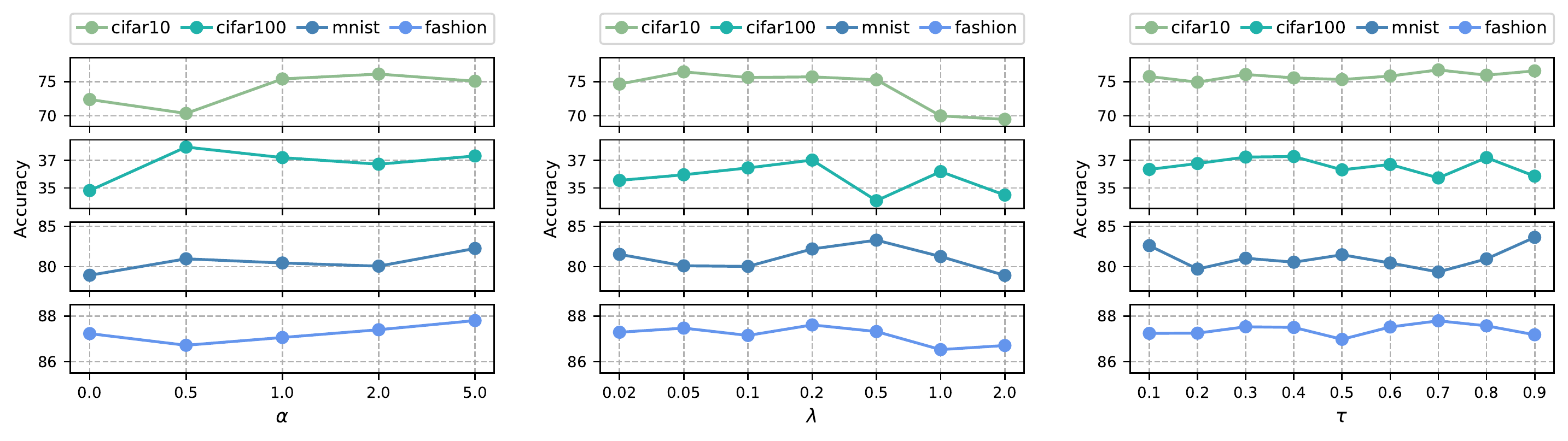}
\caption{Accuracy of setting hyper-parameters  $\alpha$, $\lambda$, and $\tau$ with different values on CIFAR-10, CIFAR-100, MNIST, and FashionMNIST.} 
\label{fig:param}  
\end{figure*}

\begin{table}[t]
\caption{Comparison of variants of our method with or without dual-branch batch normalization (DBBN). `w/o DBBN' means the conventional batch normalization with shared parameters for processing weakly and strongly augmented images is used. } \label{tab:bn}
\centering
\begin{tabular}{ >{\centering}p{0.3\linewidth}>{\centering}p{0.25\linewidth}>{\centering\arraybackslash}p{0.25\linewidth}}
\toprule
Method  & CIFAR-10 & CIFAR-100 \\ 
\midrule
\textbf{w/o DBBN}  &  74.11 & 34.22  \\ 
\midrule
\textbf{w/ DBBN}  &  \textbf{76.94}  &  \textbf{36.92} \\ 
\bottomrule
\end{tabular}
\end{table}

\subsubsection{Effectiveness of Dual-Branch Batch Normalization}
As shown in Table~\ref{tab:bn}, we study the impact of dual-branch batch normalization on four datasets.  
Thanks to the efficacy in preventing strongly augmented images from
 biasing the feature distribution, the dual-branch batch normalization can improve the classification performance consistently on all datasets.
\begin{table}[ht]
\caption{Performance of different methods in separating clean and noisy samples on FashionMNIST.}\label{tab:sep}
\centering
\setlength\tabcolsep{5.5pt}
\begin{tabular}  {l|c|c|c|c}
\toprule
\multirow{2}{*}{Method} & \multicolumn{2}{c|}{CI} & \multicolumn{2}{c}{CD}\\ \cmidrule(l){2-3}  \cmidrule(l){4-5}
 & $\eta$=0.2 & $\eta$=0.6 & $\eta$=0.2 & $\eta$=0.6 \\ \midrule
\textbf{Co-Teaching}& 65.83 & 37.86 & 56.59 & 16.50\\ 
\textbf{Co-Teaching+} & 36.58 & 19.84 & 38.13 & 17.95\\ 
\textbf{JoCoR} & 85.52 & 65.10 & 65.43 & 18.60 \\ 
\textbf{Co-Matching} & 78.21 & 65.53 & 57.51 &20.56 \\ \midrule
\textbf{Ours}  & \textbf{86.35} & \textbf{70.21}  & \textbf{86.72} & \textbf{25.34}\\ \bottomrule
\end{tabular}
\end{table}
\subsubsection{Performance of Noise Detection} 
In this subsection, we compare our method against other LLN methods in separating clean and noisy samples.
The experiments are conducted on the FashionMNIST dataset. Both class-independent and class-dependent label noises with $\eta=0.2$ or $0.6$ are used for testing, and $\rho$ is set to 0.01. The accuracy metric is used for evaluation, and the experimental results are presented in Table~\ref{tab:sep}. Our proposed method performs significantly better than other LLN methods. 

\subsubsection{Sensitivity to Hyper-parameters}
In this subsection, we discuss the sensitivity of our method against the variance of hyper-parameters, including $\alpha$, $\lambda$, and $\tau$. The experimental results are illustrated in Fig.~\ref{fig:param}. The class-independent noises are utilized for distorting image labels of four datasets CIFAR-10, CIFAR-100, MNIST, and FashionMNIST. The noise rate and imbalance ratio is set to $20\%$ and 0.01, respectively.
We can observe that the performance of our method is robust against the variance of the three hyper-parameters in most cases.

\begin{table}[t]
\caption{The impact of data augmentation. The experiments are conducted on ResNet32 (imbalance ratio $\rho$=0.01 and noise rate $\eta$=0.2). 
} \label{tab:aug}
\centering
\setlength\tabcolsep{3.5pt}
\begin{tabular}  {l|c|c}
\toprule 
\textbf{EXPERIMENTS}  & CIFAR-10 & CIFAR-100 \\ \midrule
\textbf{Single} $\longleftarrow$ \textbf{Weakly}($x_i$) & 68.82 & 29.31\\ 
\textbf{Single} $\longleftarrow$ \textbf{Strongly}($\hat{x}_i$) & 71.16 & 32.90\\ \midrule
\textbf{Dual} $\longleftarrow$ (\textbf{Weakly}, \textbf{Weakly}) ($x_i,x'_i$)& 71.88 & 33.56\\ 
\textbf{Dual} $\longleftarrow$ (\textbf{Weakly}, \textbf{Strongly}) ($x_i,\hat{x}_i$) & 76.94 & 36.92\\
\textbf{Dual} $\longleftarrow$ (\textbf{Strongly}, \textbf{Strongly}) ($\hat{x}_i,\hat{x}'_i$) & 73.13 & 34.32\\ \bottomrule
\end{tabular}
\end{table}
\subsubsection{The Impact of Data Augmentation Strategies}
Data augmentation~\cite{cubuk2018autoaugment} is a key strategy for preventing over-fitting in current deep network learning, which considerably boost performance and is widely employed in various tasks. In this subsection, we conduct experiments as Table~\ref{tab:aug} to answer why we need weak and strong augmentations. \textbf{Single} means two input images are the same while \textbf{Dual} indicates that two different instances are generated based different augmentation strategies. As can be observed that, combining weak and strong augmentations can further improve the performance.

\section{Conclusion}
In this paper, we propose an effective method to tackle the task of learning with both long-tailed distribution and label noise. 
We find that the integration of the matching between predictions on weakly and strongly augmented images is beneficial for improving the discrimination between noisy samples and clean samples of tail classes.
It is also validated that adopting dual branches to process weakly and strongly augmented images with different parameters in the batch normalization can promote the performance since it prevents the deviation of the feature distribution.
We demonstrate that the leave-noise-out regularization on noisy samples also improves the performance since it benefits in erasing the negative impact of noisy samples. 
Additionally, an online prior penalization scheme is proposed for rebalancing the training data, which can achieve the efficacy of label smoothing.
Extensive experiments on five datasets illustrate that the proposed method outperforms state-of-the-art methods.

\bibliographystyle{IEEEtran}
\bibliography{mybib}

\vfill

\end{document}